\documentclass[lettersize,journal]{IEEEtran}
\usepackage{amsmath,amsfonts}
\usepackage{algorithmic}
\usepackage{array}
\usepackage[caption=false,font=normalsize,labelfont=sf,textfont=sf]{subfig}
\usepackage{textcomp}
\usepackage{stfloats}
\usepackage{url}
\usepackage{verbatim}
\usepackage{graphicx}
\usepackage{subcaption}
\usepackage{subfig}
\usepackage{booktabs}
\usepackage{xcolor}
\usepackage{multirow}
\def\BibTeX{{\rm B\kern-.05em{\sc i\kern-.025em b}\kern-.08em
    T\kern-.1667em\lower.7ex\hbox{E}\kern-.125emX}}
\usepackage{balance}
\begin{document}
\title{Robust Metaheuristics under Uncertainty for Berth Allocation and Quay Crane Assignment: A Review}
\author{Peilan~Xu,~\IEEEmembership{Member,~IEEE}, Yang~Li, Wenjian~Luo,~\IEEEmembership{Senior Member,~IEEE}
	\thanks{This work is partly supported by the Natural Science Foundation of Jiangsu Province (*), the National Natural Science Foundation of China (*), Shenzhen Fundamental Research Program (*), Shenzhen Science and Technology Program (*).
		(\textit{Corresponding author: Peilan Xu.})}
	\thanks{Peilan Xu and Yang Li, are with School of Artificial Intelligence, Nanjing University of Information Science and Technology, Nanjing 210044, China.
		
		Wenjian Luo is with Guangdong Provincial Key Laboratory of Novel Security Intelligence Technologies, Institute of Cyberspace Security, School of Computer Science and Technology, Harbin Institute of Technology, Shenzhen 518055, Guangdong, China.}
	
	\thanks{Email: 202412492740@nuist.edu.cn, xpl@nuist.edu.cn, luowenjian@hit.edu.cn.}
}
\markboth{Journal of \LaTeX\ Class Files,~Vol.~18, No.~9, September~2020}%
{How to Use the IEEEtran \LaTeX \ Templates}

\maketitle

\begin{abstract}
The berth allocation and quay crane assignment problem (BACAP) is a representative port-terminal scheduling problem in maritime transportation and freight logistics, where vessel arrivals, berth positions, service durations, and quay-crane availability are tightly coupled. Under uncertainties such as arrival deviations, handling-time fluctuations, and resource disruptions, schedules optimized under nominal assumptions may become fragile during execution, motivating the study of robust metaheuristic optimization for BACAP in port-terminal operations. Although population-based metaheuristics have been widely used for BACAP and related port-scheduling problems, existing studies remain fragmented in their uncertainty representations, robustness criteria, search mechanisms, and empirical evaluation protocols. To the best of our knowledge, this paper provides the first focused review dedicated to robust population-based metaheuristics for BACAP under uncertainty. We first summarize uncertainty sources and information representations in BACAP, and then organize existing methods from a mechanism-oriented perspective, covering solution representation and decoding, robust evaluation and selection, robustness-guided search dynamics, and feasibility preservation and recovery. We further present a benchmark suite for uncertain BACAP to support controlled empirical comparison and report illustrative baseline results by combining representative metaheuristics with different robustness strategies. Finally, we identify open challenges related to benchmark extension, robustness-aware search design, time-adaptive robustness, and non-stationary uncertainty.
\end{abstract}

\begin{IEEEkeywords}
	Ports and terminals, optimization and control, berth allocation and quay crane assignment, robust metaheuristics.
\end{IEEEkeywords}

\section{Introduction}
\label{sec:intro}

Scheduling in port-terminal operations and related resource-intensive service systems is an important problem in intelligent optimization and combinatorial search, where service sequences, resource allocation, and queue evolution must be coordinated under uncertain arrivals, fluctuating processing times, and temporary resource disruptions \cite{li2020bi,chen2025dynamic,song2024multi}. In such settings, optimizing a nominal schedule is insufficient, because execution-time disturbances may change waiting times, resource utilization, and downstream feasibility after a schedule has been generated\cite{ben2009robust,bertsimas2011theory}. The resulting challenge is not only to improve operational efficiency under nominal assumptions, but also to preserve stability, robustness, and recoverability under uncertainty. The berth allocation and quay crane assignment problem (BACAP) provides a representative setting for studying this issue in container-terminal operations, because it couples vessel sequencing, berth-space allocation, service timing, and quay-crane deployment within a dynamic seaside operation environment\cite{bierwirth2010survey}.

BACAP is central to the initial stage of vessel service and directly influences vessel turnaround time as well as downstream terminal operations\cite{ji2022enhanced}. Berth allocation determines when and where a vessel is berthed, whereas quay crane (QC) assignment determines how handling resources are deployed and how service capacity is formed\cite{steenken2004container,park2003scheduling}. Berthing plans constrain feasible QC deployment, while QC assignment affects service duration and subsequent berth availability\cite{DAGANZO1989159}, making BACAP an intrinsically coupled scheduling problem. Under practical uncertainties such as arrival deviations, handling-time fluctuations, and equipment disruptions, schedules that perform well in deterministic settings may become fragile during execution\cite{xu2012robust}. Such disturbances may propagate through queue formation, delay accumulation, and downstream resource conflicts, thereby amplifying the execution risk of nominally high-quality schedules\cite{herroelen2005project}. This makes robustness a central consideration in method design.

\begin{figure}[htbp]
	\centering
	
	\subfloat[Annual publications on uncertain BACAP studies (2008--May 2026).%
	\label{fig:trends}]{
		\includegraphics[width=0.47\linewidth]{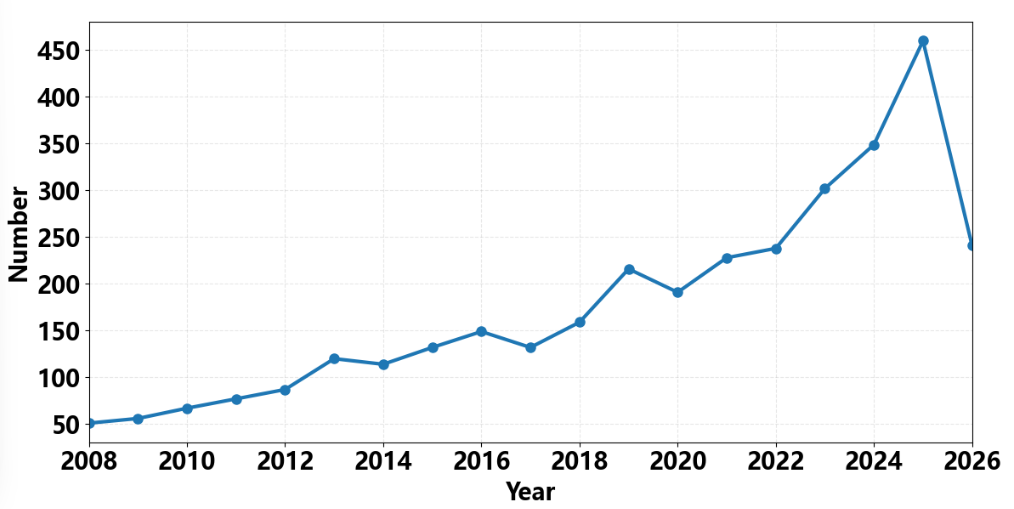}
	}%
	\hfill
	\subfloat[Algorithm distribution in uncertain BACAP literature.%
	\label{fig:bacap}]{
		\includegraphics[width=0.47\linewidth]{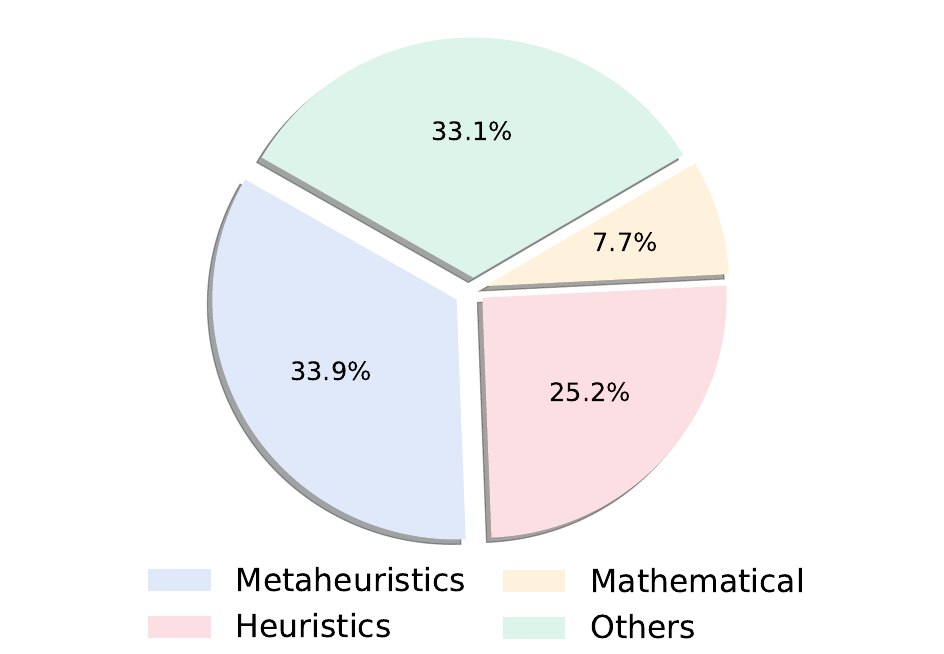}
	}
	
	\caption{BACAP literature identification and landscape analysis using a PRISMA-style screening procedure.
		(a) Annual trend of metaheuristic-based studies on uncertain BACAP.
		(b) Distribution of algorithm families in the included studies.
		Records from 2008--May 2026 were retrieved from Google Scholar using BACAP-, uncertainty-, and robustness-related keywords.
		After duplicate removal, title/abstract screening, and full-text eligibility assessment, studies were retained if they addressed uncertain berth allocation, quay crane assignment, or integrated BACAP with metaheuristic, population-based, stochastic, or robust solution methods.
		Deterministic-only, irrelevant, duplicate, incomplete, or methodologically insufficient studies were excluded.}
	\label{fig:combined}
\end{figure}

To address BACAP, existing studies have developed several methodological routes, including mathematical programming, heuristics, and metaheuristic optimization\cite{bierwirth2010survey}. Early work often relied on manual planning rules or problem-specific constructive procedures, which were suitable for small terminals with limited traffic\cite{steenken2004container}. As terminal operations became more complex, mathematical programming models and exact methods such as integer programming were introduced to obtain high-quality solutions\cite{park2003scheduling}. Although these approaches offer strong modeling expressiveness and theoretical rigor, they often face computational limitations in large-scale, highly combinatorial, or dynamically perturbed settings. Heuristics and trajectory-based metaheuristics were subsequently adopted to improve computational efficiency, but their search behavior may be sensitive to large search spaces, complex feasibility structures, and dynamic uncertainty\cite{blum2003metaheuristics}. In contrast, population-based metaheuristics, including evolutionary and swarm-based methods, maintain multiple candidate schedules simultaneously and are therefore compatible with diverse schedule exploration, uncertainty-aware evaluation, and robustness-oriented selection\cite{eiben2015introduction,kennedy1995particle}. These properties have made robust population-based metaheuristics an important research direction for uncertain BACAP and robust port scheduling \cite{chen2025dynamic,xu2025continuous}. Fig.~\ref{fig:combined} contextualizes this methodological evolution, where Fig.~\ref{fig:trends} quantifies the rapid growth of uncertain BACAP research since 2008, while Fig.~\ref{fig:bacap} reveals the dominance of metaheuristics (33.9\%) over heuristic methods (25.2\%) in current literature. 

Despite substantial progress, research on uncertain BACAP remains methodologically fragmented. Existing studies differ not only in uncertainty representations, such as probabilistic, fuzzy, interval, set-based, and scenario-based formulations, but also in the design and evaluation of robust metaheuristic approaches\cite{wang2024robust}. Although the BACAP literature includes several comprehensive reviews, these surveys are typically organized around problem variants, algorithm categories, or modeling formulations\cite{bierwirth2010survey,rodrigues2022berth}, and have not fully examined robust metaheuristics from the perspective of uncertainty-aware search design. In particular, limited attention has been paid to how key components, such as solution representation, uncertainty information for search evaluation, robustness evaluation, search dynamics, and feasibility preservation and recovery, interact in population-based approaches under uncertainty\cite{jin2005evolutionary,yang2007evolutionary,liu2021decision}. This fragmentation is reflected not only in modeling and algorithm design, but also in evaluation practices. Many existing studies on BACAP use proprietary or case-specific datasets derived from real terminal operations\cite{umang2013exact}. Although such datasets improve application realism, they also introduce substantial heterogeneity in vessel distributions, operational settings, and uncertainty representations, which makes controlled, reproducible, and transferable comparison across studies difficult.

Against this background, this paper presents a focused review of robust population-based metaheuristics for BACAP under uncertainty. The review is organized around five design dimensions of uncertainty-aware metaheuristic search: solution representation and decoding, uncertainty information for search evaluation, robust evaluation and selection, robustness-guided search dynamics, and feasibility preservation and recovery. Together, these dimensions characterize how robustness is embedded into population-based search. In addition, the paper presents a benchmark suite for uncertain BACAP and reports illustrative baseline results by combining representative metaheuristics with different robustness strategies. The benchmark study is intended to support more controlled empirical comparison and to demonstrate how different robustness paradigms can be evaluated under shared uncertainty settings. Overall, this paper aims to provide a mechanism-oriented analytical perspective on prior research, clarify methodological gaps, and support the future design of robust and reproducible metaheuristic approaches for uncertain BACAP and related port-scheduling systems.

The remainder of this paper is organized as follows. Section~\ref{sec:basic_structure} introduces the basic structure and deterministic formulation of BACAP. Section~\ref{sec:robust_metaheuristics} reviews robust population-based metaheuristics for uncertain BACAP from a mechanism-oriented perspective, including uncertainty information, robust evaluation, search dynamics, and feasibility preservation. Section~\ref{sec:sec_3} presents the benchmark suite used to evaluate solution quality and robustness under uncertain operational settings. Section~\ref{sec:sec_4} reports representative baseline results and discusses different robustness strategies. Finally, Section~\ref{sec:sec_5} concludes the paper and outlines future research directions.
\section{Basic Structure of the Berth Allocation and Quay Crane Assignment Problem}
\label{sec:basic_structure}

BACAP concerns the coordinated allocation of limited quay space and QC resources to arriving vessels in order to improve service efficiency at container terminals. In its basic form, the problem determines where and when each vessel should berth, as well as how QC resources should be assigned during its service period, so as to optimize operational objectives such as vessel waiting time, turnaround time, service completion time, or overall operating cost.

Rather than forming two separable subproblems, BACAP defines a tightly coupled spatiotemporal scheduling structure. Berth allocation determines the service start time and spatial position of each vessel along the quay, whereas QC assignment determines the service capacity available to that vessel and directly affects handling duration and departure time. The completion time of one vessel further influences berth release and the feasibility of subsequent allocations for other vessels. This mutual dependence prevents berth-related and QC-related decisions from being treated independently in realistic scheduling settings.

From a modeling perspective, BACAP involves berth-related decisions, including vessel berthing positions, service start times, and service completion times, as well as QC-related decisions, including the number of QCs assigned to each vessel, their temporal usage, and, when explicitly modeled, their spatial deployment along the quay. These decisions are constrained by quay boundaries, berth non-overlap, vessel arrival-time restrictions, QC assignment bounds, QC interference and safety-distance requirements, and the exclusivity of each QC's assignment at a given time. Depending on the terminal setting, additional constraints may also be introduced to reflect service windows, tidal restrictions, service priorities, draft limitations, or terminal-specific operating rules.

Table~\ref{tab:basic_bacap_notation} summarizes the main notation used in a representative deterministic BACAP formulation.

\begin{table}[t]
	\caption{Parameters and decision variables for a representative deterministic BACAP formulation.}
	\label{tab:basic_bacap_notation}
	\centering
	\small
	\renewcommand{\arraystretch}{1.12}
	\begin{tabular}{p{0.30\columnwidth} p{0.60\columnwidth}}
		\toprule
		\multicolumn{2}{c}{\textbf{Sets and indices}} \\
		\midrule
		$\mathcal{V}=\{1,\dots,n\}$ & Set of vessel indices \\
		$\mathcal{Q}=\{1,\dots,m\}$ & Set of QC indices \\
		$\mathcal{T}=\{0,\dots,H-1\}$ & Set of discrete time periods \\
		$i,j \in \mathcal{V}$ & Vessel indices \\
		$k,\ell \in \mathcal{Q}$ & QC indices \\
		$t \in \mathcal{T}$ & Time-period index \\
		\midrule
		\multicolumn{2}{c}{\textbf{Parameters}} \\
		\midrule
		$a_i\in\{0,\dots,H\}$ & Nominal arrival time of vessel $i$ \\
		$l_i\in(0,L]$ & Length of vessel $i$ \\
		$p_i>0$ & Handling workload of vessel $i$ \\
		$\mu>0$ & Handling productivity of one QC per time period \\
		$[\underline{t}_g,\overline{t}_g]\subseteq[0,H]$ & The g-th tidal time window \\
		$L>0$ & Total quay length \\
		$b\ge 0$ & Minimum safety distance between simultaneously operating QCs \\
		$1\le \underline{q}_i\le \overline{q}_i\le m$ & Minimum and maximum numbers of QCs assignable to vessel $i$ \\
		$M>0$ & A sufficiently large constant \\
		\midrule
		\multicolumn{2}{c}{\textbf{Decision and auxiliary variables} $\mathbf{X}$} \\
		\midrule
		$x_i\in[0,L-l_i]$ & Berthing position of vessel $i$ \\
		$s_i\in\{0,\dots,H\}$ & Service start time of vessel $i$ \\
		$d_i\in\{0,\dots,H\}$ & Completion time of vessel $i$ \\
		$z_{i,t}\in\{0,1\}$ & Equals 1 if vessel $i$ is under service during time period $t$, and 0 otherwise \\
		$y_{i,k,t}\in\{0,1\}$ & Equals 1 if QC $k$ serves vessel $i$ during time period $t$, and 0 otherwise \\
		$q_{i,t}\in\{0,\dots,m\}$ & Number of QCs assigned to vessel $i$ during time period $t$ \\
		$\alpha_{i,j}\in\{0,1\}$ & Equals 1 if vessel $i$ is spatially before vessel $j$, and 0 otherwise \\
		$\beta_{i,j}\in\{0,1\}$ & Equals 1 if vessel $i$ is temporally before vessel $j$, and 0 otherwise \\
		$\gamma_{i,j}\in\{0,1\}$ & Equals 1 if vessels $i$ and $j$ are separated in space, and 0 if they are separated in time \\
		$c_{i,k,t}\in[0,L]$ & Position of QC $k$ when assigned to vessel $i$ during time period $t$ \\
		\bottomrule
	\end{tabular}
\end{table}

Based on the notation summarized in Table~\ref{tab:basic_bacap_notation}, the following formulation captures the basic deterministic structure of BACAP.

\begin{equation}
	\label{eq:objective}
	\min \sum_{i \in \mathcal{V}} (d_i-a_i)
\end{equation}

\begin{equation}
	\label{eq:start_time}
	s_i \ge a_i, \quad \forall i \in \mathcal{V}
\end{equation}

\begin{equation}
	\label{eq:TideWindow}
	\underline{t}_g \le s_i \le \overline{t}_g,\quad \forall i \in \mathcal{V}
\end{equation}

\begin{equation}
	\label{eq:completion_after_start}
	s_i \le d_i,\quad \forall i \in \mathcal{V}
\end{equation}

\begin{equation}
	\label{eq:berth_conflict1}
	x_i + l_i \le x_j + M(1-\alpha_{i,j}) + M(1-\gamma_{i,j}),
	\quad \forall i,j \in \mathcal{V},\, i<j
\end{equation}

\begin{equation}
	\label{eq:berth_conflict2}
	x_j + l_j \le x_i + M\alpha_{i,j} + M(1-\gamma_{i,j}),
	\quad \forall i,j \in \mathcal{V},\, i<j
\end{equation}

\begin{equation}
	\label{eq:berth_time_1}
	s_j \ge d_i - M(1-\beta_{i,j}) - M\gamma_{i,j},
	\quad \forall i,j \in \mathcal{V},\, i<j
\end{equation}

\begin{equation}
	\label{eq:berth_time_2}
	s_i \ge d_j - M\beta_{i,j} - M\gamma_{i,j},
	\quad \forall i,j \in \mathcal{V},\, i<j
\end{equation}

\begin{equation}
	\label{eq:service_link_start}
	s_i \le t + M(1-z_{i,t}),\quad \forall i\in\mathcal{V},\, \forall t\in\mathcal{T}
\end{equation}

\begin{equation}
	\label{eq:service_link_end}
	d_i \ge t+1 - M(1-z_{i,t}),\quad \forall i\in\mathcal{V},\, \forall t\in\mathcal{T}
\end{equation}

\begin{equation}
	\label{eq:qca_service_link}
	y_{i,k,t} \le z_{i,t},\quad \forall i\in\mathcal{V},\, \forall k\in\mathcal{Q},\, \forall t\in\mathcal{T}
\end{equation}

\begin{equation}
	\label{eq:qc_number_definition}
	q_{i,t}=\sum_{k\in\mathcal{Q}}y_{i,k,t},\quad \forall i\in\mathcal{V},\, \forall t\in\mathcal{T}
\end{equation}

\begin{equation}
	\label{eq:crane_number}
	\underline{q}_i z_{i,t}
	\le q_{i,t}
	\le \overline{q}_i z_{i,t},
	\quad \forall i\in\mathcal{V},\, \forall t\in\mathcal{T}
\end{equation}

\begin{equation}
	\label{eq:v_c_t}
	\sum_{i \in \mathcal{V}} y_{i,k,t} \le 1,
	\quad \forall k \in \mathcal{Q},\, \forall t\in\mathcal{T}
\end{equation}

\begin{equation}
	\label{eq:workload_completion}
	\sum_{t\in\mathcal{T}}\sum_{k\in\mathcal{Q}}\mu y_{i,k,t}
	\ge p_i,
	\quad \forall i\in\mathcal{V}
\end{equation}

\begin{equation}
	\label{eq:crane_position_link_1}
	c_{i,k,t}\ge x_i - M(1-y_{i,k,t}),
	\quad \forall i\in \mathcal{V},\, \forall k\in \mathcal{Q},\, \forall t\in\mathcal{T}
\end{equation}

\begin{equation}
	\label{eq:crane_position_link_2}
	c_{i,k,t}\le x_i+l_i+M(1-y_{i,k,t}),
	\quad \forall i\in\mathcal{V},\, \forall k\in \mathcal{Q},\, \forall t\in\mathcal{T}
\end{equation}

\begin{equation}
	\label{eq:crane_distance_1}
	\begin{aligned}
		&|c_{i,k,t}-c_{j,\ell,t}| \ge b - M\bigl(2-y_{i,k,t}-y_{j,\ell,t}\bigr),\\
		&\qquad \forall i,j\in\mathcal{V},\; \forall k,\ell\in\mathcal{Q},\; (i,k)\neq(j,\ell),\; \forall t\in\mathcal{T}.
	\end{aligned}
\end{equation}

The objective in \eqref{eq:objective} minimizes the nominal total port time across all vessels. The domains of the parameters and variables are specified in Table~\ref{tab:basic_bacap_notation}. Constraint~\eqref{eq:start_time} ensures that service cannot start before the vessel arrives. Constraint~\eqref{eq:TideWindow} imposes the service-start window, which can represent a simplified tidal restriction. Constraint~\eqref{eq:completion_after_start} ensures that completion does not occur before service starts.

Constraints~\eqref{eq:berth_conflict1}--\eqref{eq:berth_time_2} describe the berth--time non-overlap relationship between each pair of vessels. The variable $\gamma_{i,j}$ determines whether two vessels are separated spatially or temporally. When $\gamma_{i,j}=1$, constraints~\eqref{eq:berth_conflict1} and~\eqref{eq:berth_conflict2} enforce spatial separation according to the berth order indicated by $\alpha_{i,j}$. When $\gamma_{i,j}=0$, constraints~\eqref{eq:berth_time_1} and~\eqref{eq:berth_time_2} enforce temporal separation according to the service order indicated by $\beta_{i,j}$.

Constraints~\eqref{eq:service_link_start} and~\eqref{eq:service_link_end} link the service-state variable $z_{i,t}$ to the service interval of vessel $i$. Constraint~\eqref{eq:qca_service_link} allows a QC to be assigned only to a vessel under service. Constraint~\eqref{eq:qc_number_definition} defines the time-dependent number of QCs assigned to each vessel. Constraint~\eqref{eq:crane_number} activates the lower and upper QC assignment bounds during the service period. Constraint~\eqref{eq:v_c_t} ensures that each QC serves at most one vessel at any time. Constraint~\eqref{eq:workload_completion} ensures that the allocated QC capacity is sufficient to complete the handling workload of each vessel.

Constraints~\eqref{eq:crane_position_link_1} and~\eqref{eq:crane_position_link_2} restrict an assigned QC to operate within the spatial interval of the corresponding vessel. Constraint~\eqref{eq:crane_distance_1} maintains the minimum safety distance between simultaneously operating QCs. The absolute-value relation in \eqref{eq:crane_distance_1} can be linearized with auxiliary ordering variables. The service-state relation in \eqref{eq:service_link_start} and \eqref{eq:service_link_end} can be strengthened with reverse implications or interval-continuity constraints when an exact time-indexed implementation is required. Multiple service or tidal windows can be handled by adding window-selection variables.

The main difficulty of BACAP lies not only in the size of the combinatorial search space, but also in the strong coupling among multiple resources and the strict feasibility requirements imposed by real terminal operations. A locally favorable adjustment in berth position or QC deployment may induce cascading effects on downstream vessel service and resource usage. Consequently, even before uncertainty is explicitly introduced, BACAP already exhibits structural properties that make solution representation and decoding, feasibility preservation and recovery, and adaptive search behavior central to effective algorithm design.
\section{Robust Population-Based Metaheuristics for BACAP under Uncertainty}
\label{sec:robust_metaheuristics}
In this section, we review robust population-based metaheuristics for BACAP from a mechanism-oriented perspective, focusing on five interacting components: solution representation and decoding, uncertainty information for search evaluation, robust evaluation, ranking, and selection, robustness-guided search dynamics, and feasibility preservation and recovery.

Population-based metaheuristics are well suited for uncertain BACAP because they maintain a set of candidate schedules, enabling exploration of alternative berth--QC allocations, evaluation under uncertain conditions, preservation of diverse robustness profiles, and integration of repair mechanisms. This population-level structure provides an algorithmic basis for combining uncertainty representation, robust evaluation, search adaptation, and feasibility control \cite{eiben2015introduction,kennedy1995particle,bonabeau1999swarm}.

\begin{figure*}[t]
	\centering
	\small
	\includegraphics[width=\textwidth]{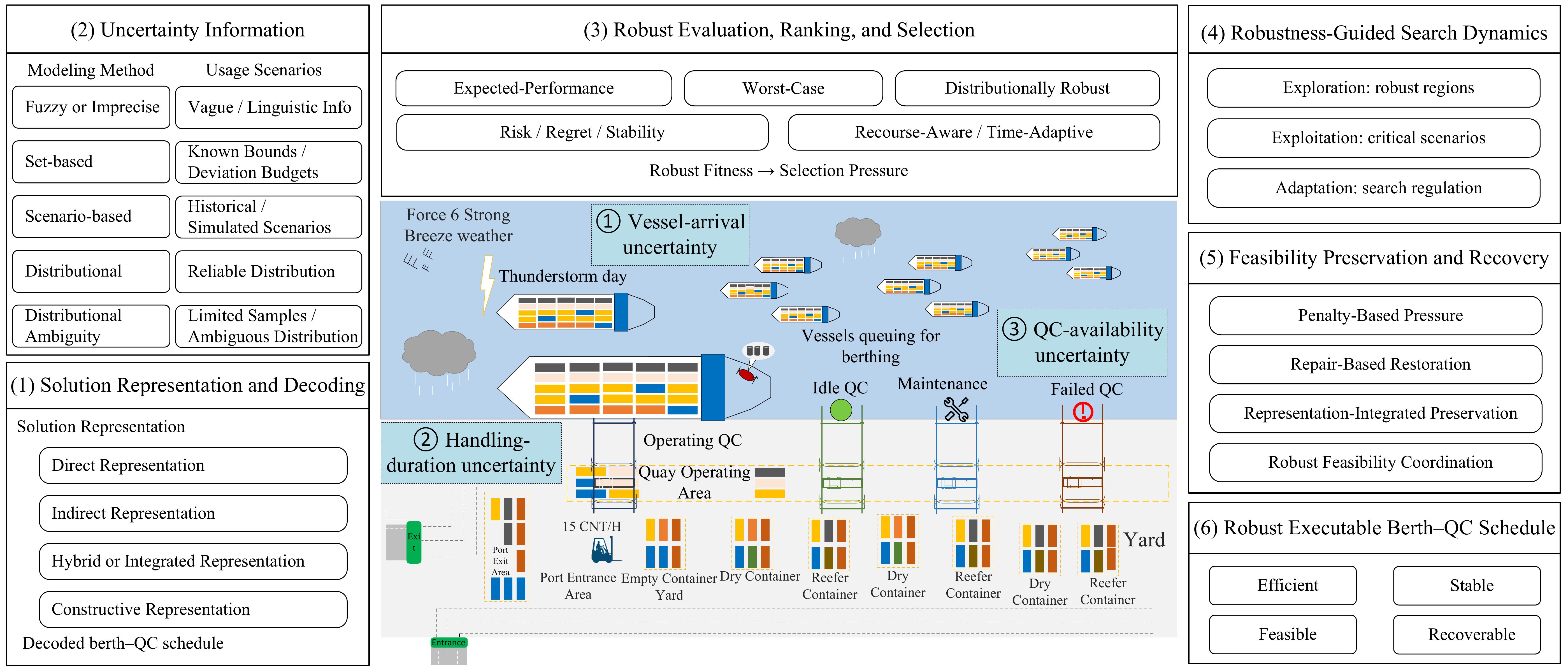}
	\caption{Conceptual framework of robust population-based metaheuristics for uncertain BACAP. The central port scenario illustrates vessel-arrival uncertainty, handling-duration uncertainty, and QC-availability uncertainty, while the surrounding modules summarize how representation, uncertainty information, robust evaluation, search dynamics, and feasibility recovery jointly shape robust executable berth--QC schedules.}
	\label{fig:robust_metaheuristic_framework}
\end{figure*}

In uncertain BACAP, discrete scheduling decisions, limited quay and QC resources, coupled constraints, and execution-time disturbances jointly shape schedule performance. Uncertainty alters not only objective values but also solution ranking, search landscapes, and feasibility. Therefore, robustness should be embedded into the metaheuristic search process through representation, evaluation, search dynamics, and feasibility recovery, rather than treated only as a post-processing criterion. Figure~\ref{fig:robust_metaheuristic_framework} illustrates this mechanism-oriented framework.

\subsection{Solution Representation and Decoding}
\label{subsec:representation_decoding}

Representation design plays a fundamental role in solving BACAP instances with population-based metaheuristics such as genetic algorithms (GAs), particle swarm optimization (PSO), and ant colony optimization (ACO). Beyond specifying candidate schedules, representation design serves as a critical interface between problem structure and search dynamics, directly influencing feasibility preservation, search-landscape characteristics, and convergence behavior. For BACAP, an individual in a population does not represent a single decision, but a coupled scheduling structure involving vessel order, berthing position, berthing start time, service completion time, QC assignment, and, in some formulations, time-indexed QC deployment. The decoder maps this encoded structure into an executable berth--QC schedule under quay-boundary, non-overlap, arrival-time, QC-capacity, and safety-distance constraints. Representation and decoding should therefore be considered jointly: the representation determines what scheduling information is represented and explored, while the decoder determines how the searched information becomes an operational schedule \cite{bean1994genetic,cheng1996tutorial,londe2025biased,aickelin2004indirect}.

Existing BACAP representations can be organized according to the relationship between the encoded individual and the decoded schedule. We distinguish four representation paradigms: direct representation, indirect representation, hybrid or integrated representation, and constructive representation. Integer-based, real-valued, random-key, matrix-based, and sequence-based encodings are treated as concrete encoding types under these paradigms.

\textit{Direct representation.} Direct representation explicitly encodes scheduling decisions themselves \cite{yu2019genetic,ji2022enhanced}. For example, integer-based encodings may directly use discrete identifiers to represent vessel service orders, berth indices, berthing positions, QC numbers, or vessel--QC assignment relations, while matrix-based encodings may directly describe vessel--time, berth--time, or vessel--QC relations \cite{cheimanoff2022exact,al2024improved,zhao2026dhmoga,tsai2015novel,wang2009multi,tengecha2022efficient}. In this paradigm, decoding is relatively direct: the encoded variables are mapped to schedule components, and the decoder mainly checks, completes, or repairs feasibility. Its advantage is operational interpretability, since each gene or matrix element has a clear scheduling meaning. However, this directness also makes the representation sensitive to uncertain arrivals and handling durations. A schedule encoded under nominal conditions may violate berth conflicts, service windows, or QC-capacity constraints when the realized operating condition changes. Thus, direct representation often requires realization-based feasibility checking or repair rather than a purely deterministic decoder.

\textit{Indirect representation.} Indirect representation does not encode the complete schedule itself, but encodes priority values, ordering keys, or scheduling preferences that are translated into a schedule by a decoder. Real-valued and random-key encodings are typical examples: the encoded values are first sorted or interpreted as priorities, and the decoder then constructs vessel sequences, berth allocations, or QC assignments according to these priorities \cite{correcher2017biased,chaves2024adaptive}. Compared with direct representation, this design often provides a smoother search space and allows standard variation operators to modify priorities without immediately destroying the whole schedule. The main limitation is that search performance depends heavily on decoder design \cite{londe2024early,londe2025biased}. In uncertain BACAP, the decoder cannot simply transform priority values into a nominal feasible schedule; it must also examine whether the decoded schedule remains feasible or recoverable under arrival deviations, handling-time fluctuations, and changing QC availability \cite{shang2016robust}.

\textit{Hybrid or integrated representation.} Hybrid or integrated representation combines direct and indirect components to describe the multi-level structure of BACAP. For instance, one part of an individual may encode vessel priorities or service sequences, while another part encodes berth allocation, timing adjustment, QC numbers, or vessel--QC matching. Such designs are useful because BACAP is not merely a sequencing problem. Berth allocation constrains feasible QC deployment, while QC assignment affects handling duration and berth release. Hybrid and hierarchical encodings have therefore been introduced to capture dependencies among vessel sequencing, berth allocation, and QC assignment \cite{tan2025integrated,wu2025ternary}. Their strength lies in expressiveness, especially for integrated berth--QC decisions, but the resulting representation is often larger and requires more problem-specific operators. Under uncertainty, richer encoded information may preserve more adjustment flexibility, but it may also increase the risk of constructing a detailed nominal schedule that is difficult to adapt when different encoded layers respond differently to arrival or handling-time perturbations.

\textit{Constructive representation.} Constructive representation generates a schedule incrementally rather than encoding a complete solution in advance \cite{bremer2021ant,zhang2023minimizing}. This form is common in ACO-based BACAP methods, where pheromone trails and heuristic information guide the stepwise selection of vessels, berthing positions, or QC assignments \cite{tong1999ant,yu2023enhanced}. In this paradigm, decoding is embedded in the construction process: after each partial decision, the algorithm can check berth availability, QC capacity, or conflict constraints before adding the next decision. This makes constructive representation attractive for strongly constrained BACAP because feasibility control can be introduced during schedule generation. However, under uncertainty, the heuristic information used for construction is often derived from nominal or sampled conditions. A locally feasible construction step may still lead to a fragile schedule if future arrivals or service durations deviate from the assumed condition. Robust constructive methods therefore require uncertainty-aware heuristic information, robust pheromone updating, or look-ahead feasibility evaluation \cite{paprocka2020method,zhang2024dynamic}.

This taxonomy highlights that the difference among representation strategies lies not only in the data type of the search object or encoding, but also in the way an encoded individual is converted into an executable schedule. Direct representations emphasize interpretability, indirect representations emphasize search smoothness, hybrid representations emphasize coupled decision expressiveness, and constructive representations emphasize stepwise schedule generation. For robust BACAP, the central issue is whether the representation--decoder pair can support uncertainty-aware evaluation and feasibility recovery.

\begin{table}[t]
	\centering
	\caption{Taxonomy of representation and encoding strategies for BACAP.}
	\label{tab:representation_decoding_taxonomy}
	\footnotesize
	\renewcommand{\arraystretch}{1.15}
	\begin{tabular}{p{2.7cm} p{5cm}}
		\toprule
		\textbf{Representation paradigm} 
		& \textbf{Representative references} \\
		\midrule
		Direct representation 
		& \cite{wang2024robust}\cite{yu2019genetic}\cite{ji2022enhanced}\cite{cheimanoff2022exact} \cite{al2024improved}\cite{zhao2026dhmoga} \cite{tsai2015novel}\cite{tengecha2022efficient}\cite{shang2016robust}\cite{wu2025ternary}\cite{hu2022improved} \cite{han2006algorithm}  \cite{rodriguez2014genetic}\cite{arango2013genetic} \\
		\midrule
		Indirect representation 
		& \cite{correcher2017biased}\cite{chaves2024adaptive} \cite{lalla2014biased} \cite{correcher2024berth}\\
		\midrule
		Hybrid or integrated representation 
		& \cite{tan2025integrated} \cite{zheng2023integrated}\cite{hu2015multi}\cite{de2020hybrid}\cite{yu2024integrated}\cite{jo2025hierarchical}\cite{wang2024proactive}\\
		\midrule
		Constructive representation 
		&  \cite{tong1999ant} \cite{yu2023enhanced}\cite{cheong2008multi}\cite{wang2022adaptive}\cite{han2010proactive}\cite{azza2014ant}\cite{yu2011application}\\
		\bottomrule
	\end{tabular}
\end{table}

\subsection{Uncertainty Information for Search Evaluation}
\label{subsec:uncertainty_search_evaluation}

Uncertainty in BACAP arises from several operational sources. Arrival-related uncertainty may be caused by weather conditions, maritime traffic, upstream port delays, and navigation variability, which may shift the actual arrival time away from the planned schedule \cite{yan2019dynamic,SCHEPLER2019363}. Handling-related uncertainty reflects fluctuations in container volume, QC productivity, equipment status, labor efficiency, and weather-induced operational delays \cite{krimi2020modelling,HE2021101252}. Resource and operational disruptions further include QC unavailability, feeder-vessel uncertainty, renewable-energy intermittency, tidal restrictions, and unplanned vessel arrivals \cite{IRIS2021102445,ji2022hybrid,zheng2023integrated}. These uncertainties affect berth availability, service completion, QC deployment, and downstream feasibility, and therefore shape how candidate schedules should be evaluated during search.

The representation of uncertainty depends on the information available for these sources. Some studies rely on imprecise expert estimates; some use admissible ranges or deviation budgets; some construct finite scenarios from historical records, simulation, or sampling; some estimate probability distributions; and some further describe uncertainty over the probability distribution itself. Accordingly, this subsection classifies uncertainty information according to how it is transformed into search evaluation: fuzzy or imprecise representation, set-based representation, scenario-based representation, distributional representation, and distributional-ambiguity representation. Data-driven and predictive techniques are treated as information-generation tools rather than independent representation categories, because they may support scenarios, distributions, uncertainty sets, or ambiguity sets.

\textit{Fuzzy or imprecise representation.} When statistical observations are limited, uncertain parameters can be described by fuzzy or imprecise information \cite{zimmermann2011fuzzy}. In BAP/BACAP, triangular fuzzy numbers, fuzzy intervals, and related fuzzy representations have been used to model ambiguous vessel arrival times or vague operational information \cite{segura2017fully,gutierrez2019fully,lujan2021fuzzy}. Such information is usually transformed into evaluable quantities through defuzzification, $\alpha$-cuts, fuzzy comparison, or possibility-based feasibility assessment. For population-based search, this representation allows candidate schedules to be evaluated when probabilistic data are unavailable, but the resulting ranking depends on the chosen fuzzy interpretation. Hybrid fuzzy--robust or fuzzy--stochastic models further combine fuzzy information with robustness or chance constraints to improve modeling flexibility.

\textit{Set-based representation.} When reliable probability distributions are unavailable but operational bounds can be estimated, uncertainty is often represented by a set of possible realizations \cite{ben2002robust,bertsimas2004price}. Interval sets describe lower and upper bounds of uncertain arrivals or handling times, while budgeted or multi-constrained uncertainty sets restrict the total magnitude of simultaneous deviations \cite{rodrigues2021exact,kolley2023robust}. For example, arrival-time deviations can be bounded by an uncertainty set, multi-source disturbances can be controlled by a budgeted set, and renewable-energy or equipment-related uncertainty can be described through constrained deviation regions \cite{CHARGUI2023106251}. In population-based search, set-based representation usually leads to robust or worst-case evaluation over the admissible region. It can improve feasibility protection but may guide selection toward conservative schedules.

\textit{Scenario-based representation.} Scenario-based representation uses a finite set of realizations to describe uncertain operating conditions. These scenarios may be obtained from historical records, simulation, sampling from assumed distributions, predictive models, or benchmark generators \cite{li2021review,chou2023problem}. In BACAP, scenario-based stochastic models have been used to approximate uncertain vessel arrivals, handling durations, container-volume fluctuations, and multi-source disruptions \cite{ma2021stochastic,wu2025ternary,tan2020berth,zheng2024integrated}. During metaheuristic search, each individual is decoded into a schedule and evaluated across the scenario set. The resulting performance values can then be aggregated by mean value, worst-case value, regret, feasibility frequency, or robustness-oriented indicators. Fixed scenario sets provide stable comparisons among individuals, whereas resampling or rolling scenario generation can reduce overfitting to a limited scenario pool at the cost of noisier fitness estimates.

\textit{Distributional representation.} When sufficient historical data or reliable statistical assumptions are available, uncertain parameters can be modeled as random variables. Arrival delays, handling durations, maritime-traffic effects, vessel-performance deviations, and equipment-related disturbances have been represented by a variety of distributions. For vessel arrival times, Poisson or normal distributions are frequently used \cite{yan2019dynamic,SCHEPLER2019363,umang2017real}; for handling durations and operational disturbances, exponential, uniform, or empirically fitted distributions are common \cite{BUDIPRIYANTO2017127,TANG2024110038,JIA2020174}. Markov processes can also be used when weather, labor efficiency, or QC availability evolves across time states \cite{URSAVAS2016380}. In population-based metaheuristics, distributional information enters evaluation through sampling, simulation, or analytical expectation. It supports expected-performance evaluation, risk measures, variance-based stability assessment, and chance-type feasibility measures. Its reliability depends on the quality of distribution estimation and may degrade when historical data do not represent future operating conditions.

\textit{Distributional-ambiguity representation.} When historical samples are available but insufficient to identify a single reliable probability distribution, uncertainty can be represented by an ambiguity set, i.e., a set of plausible probability distributions. Unlike a conventional uncertainty set, which contains possible parameter realizations, an ambiguity set contains possible distributions of those realizations. BACAP studies have used moment information \cite{delage2010distributionally,nie2023distributionally} and Wasserstein-distance-based ambiguity sets \cite{mohajerin2018data,gao2023distributionally} to model uncertain handling times and other distributionally uncertain parameters \cite{AGRA20221,rodrigues2024handling,wang2025optimizing,wang2024distributionally,tang2023distributionally}. In search evaluation, this representation provides the information basis for distributionally robust optimization: a candidate schedule is assessed against the worst expected performance over all distributions in the ambiguity set. This can improve robustness against distributional misspecification, but it also increases evaluation complexity because fitness computation may involve an inner worst-distribution problem.

These representations differ in the type and reliability of uncertainty information they require, but they share a common algorithmic role: they define how uncertain operating conditions are transformed into fitness evaluation. Fuzzy and set-based representations are suitable when statistical information is weak; scenario-based representations provide a direct multi-realization evaluation interface; distributional representations support expectation and risk-based evaluation; and ambiguity sets support distributionally robust evaluation. Once embedded into population-based metaheuristics, these representations change solution ranking, selection pressure, evaluation cost, and the shape of the search landscape.

\begin{table}[t]
	\centering
	\caption{Taxonomy of uncertainty information representations for BACAP search evaluation.}
	\label{tab:uncertainty_representation_taxonomy}
	\footnotesize
	\renewcommand{\arraystretch}{1.15}
	\begin{tabular}{p{4.4cm} p{3.4cm}}
		\toprule
		\textbf{Representation type}
		& \textbf{Representative references} \\
		\midrule
		Fuzzy or imprecise representation
		&\cite{segura2017fully}\cite{gutierrez2019fully}\cite{lujan2021fuzzy}\cite{perez2022multi}\\
		\midrule
		Set-based representation
		&\cite{rodrigues2021exact}\cite{kolley2023robust}\cite{CHARGUI2023106251}\cite{liu2020two}\cite{xiang2021almost}\\
		\midrule
		Scenario-based representation
		& \cite{ji2022enhanced}\cite{wu2025ternary}\cite{ma2021stochastic}\cite{tan2020berth}\cite{zheng2024integrated}\cite{jmse12091541}\cite{jiang2022integrated}\\
		\midrule
		Distributional representation
		& \cite{han2010proactive}\cite{yan2019dynamic}\cite{SCHEPLER2019363}\cite{umang2017real}\cite{BUDIPRIYANTO2017127}\cite{JIA2020174}\cite{URSAVAS2016380}
		\cite{ZHOU2008161} \\
		\midrule
		Distributional-ambiguity representation
		& \cite{AGRA20221}\cite{rodrigues2024handling}\cite{wang2025optimizing}\cite{wang2024distributionally}
		\cite{tang2023distributionally}\\
		\bottomrule
	\end{tabular}
\end{table}

\subsection{Robust Evaluation, Ranking, and Selection}
\label{subsec:robust_evaluation_selection}

After candidate schedules are generated through population update, variation, neighborhood search, or swarm interaction, the algorithm must evaluate and rank them before selection. In deterministic BACAP, this ranking is usually based on a nominal objective value, such as total port time, waiting time, delay penalty, or operating cost. Under uncertainty, however, the same decoded schedule may exhibit different performance under different vessel arrivals, handling durations, QC availability states, or operational disruptions. A schedule that is attractive under nominal conditions may therefore become fragile, showing large degradation or infeasibility under disturbed realizations. Robust evaluation addresses this issue by converting uncertainty-dependent performance into a ranking signal that directly guides population selection.

Let $\mathbf{x}$ denote a decoded berth--QC schedule and $\omega$ denote an uncertain realization. The performance of $\mathbf{x}$ under $\omega$ can be written as $F(\mathbf{x},\omega)$, where $F$ may include service cost, delay, port time, penalty terms, or other operational indicators. Robust evaluation defines an aggregation rule $\Phi(\mathbf{x})$ over scenarios, distributions, uncertainty sets, ambiguity sets, or recourse processes. This value is then used for ranking and selection. In this setting, robustness is not only a post-processing criterion; it changes the selection pressure of the metaheuristic search. Existing mechanisms can be organized according to two aspects: how uncertainty-dependent performance is aggregated for a fixed schedule, and whether the schedule is evaluated as a fixed plan or as a plan with recourse and time-adaptive adjustment.

\textit{Expected-performance evaluation.} Expected-performance evaluation ranks each candidate schedule by its average or expected performance under uncertain realizations. When a finite scenario set $\Omega=\{\omega_1,\ldots,\omega_K\}$ is available, a typical evaluation form is $\Phi_{\mathrm{E}}(\mathbf{x})=\sum_{k=1}^{K}p_k F(\mathbf{x},\omega_k)$, where $p_k$ is the scenario weight and $\sum_{k=1}^{K}p_k=1$. This criterion favors schedules with good overall efficiency across representative scenarios and is usually less conservative than worst-case evaluation. In population-based search, it creates selection pressure toward solutions with strong average performance and avoids excessive protection against rare disruptions. Its weakness is that severe tail events or low-probability infeasible realizations may be hidden by the average. In BACAP, expected-performance evaluation is therefore suitable as an efficiency-oriented robustness baseline, especially when scenario probabilities or empirical frequencies are reliable \cite{zhen2024primal}.

\textit{Worst-case evaluation.} Worst-case evaluation ranks a schedule according to its poorest performance over a scenario set or uncertainty set, e.g., $\Phi_{\mathrm{W}}(\mathbf{x})=\max_{\omega\in\Omega}F(\mathbf{x},\omega)$ or $\Phi_{\mathrm{W}}(\mathbf{x})=\max_{\mathbf{u}\in\mathcal{U}}F(\mathbf{x},\mathbf{u})$. This mechanism corresponds to the min--max robust optimization (MMRO) principle and emphasizes protection under adverse realizations \cite{ben2009robust,bertsimas2011theory}. In BACAP, interval-based uncertainty sets, budgeted uncertainty sets, and scenario-based worst-case formulations have been used to regulate conservatism and protect schedules against uncertain arrivals, handling durations, and operational disturbances \cite{shang2016robust,nourmohammadzadeh2022robust,liu2020two,rodrigues2021exact}. In metaheuristic selection, worst-case evaluation favors conservative schedules with larger operational buffers or more stable resource allocation. This can improve reliability, but may sacrifice average efficiency and reduce the diversity of high-performing alternatives.

\textit{Risk-, regret-, and stability-aware evaluation.} Between expected and worst-case evaluation, several criteria measure robustness through reliability, relative loss, or sensitivity. Risk-aware evaluation considers violation probability, tail performance, conditional value-at-risk (CVaR), or survival probability, and thus favors schedules with low probability of severe disruption \cite{royset2025risk,kuccukyavuz2022chance}. Regret-based evaluation ranks a schedule by its loss relative to the best schedule in each realization, encouraging solutions that remain competitive across different operating conditions \cite{groetzner2022multiobjective,baak2025robust}. Stability-aware evaluation measures performance variation, such as variance or sensitivity to perturbations, and favors schedules whose quality changes smoothly under uncertainty. This idea is related to the broader dynamic-robustness perspective discussed in robust optimization over time \cite{yazdani2023robust}. These criteria induce selection pressure toward reliable and stable schedules rather than merely average-efficient or worst-case-protected ones.

\textit{Distributionally robust evaluation.} Distributionally robust evaluation is used when the true probability distribution is uncertain. Instead of evaluating a schedule under one assumed distribution, it evaluates the worst expected performance over an ambiguity set $\mathcal{P}$ of plausible distributions \cite{kuhn2025distributionally}, i.e., $\Phi_{\mathrm{DRO}}(\mathbf{x})=\sup_{\mathbb{P}\in\mathcal{P}}\mathbb{E}_{\mathbb{P}}[F(\mathbf{x},\omega)]$. In BACAP, ambiguity sets based on moment information \cite{delage2010distributionally,nie2023distributionally} or Wasserstein distance \cite{mohajerin2018data,gao2023distributionally} have been used to model distributional uncertainty in handling times and other uncertain parameters \cite{AGRA20221,rodrigues2024handling,wang2025optimizing}. This criterion changes ranking by favoring schedules that are insensitive to distribution misspecification, rather than schedules optimized for a single empirical distribution. Its main limitation is computational cost, because evaluating each candidate may require solving or approximating an inner worst-distribution problem.

\textit{Recourse-aware and time-adaptive evaluation.} The preceding criteria mainly evaluate a decoded schedule as a fixed plan. In real port operations, however, part of the decision may be adjusted after uncertainty is realized. Two-stage robust optimization (TSRO) captures this structure by distinguishing first-stage plans from second-stage recourse decisions. In BACAP, first-stage decisions may include vessel priorities, planned berthing positions, or initial QC allocation, while second-stage decisions may adjust service start times, QC deployment, or recovery actions after actual arrivals and handling conditions are observed. Such evaluation ranks schedules not only by their initial performance, but also by their recoverability, recourse cost, and post-disruption feasibility \cite{IRIS2019365,qu2023two,wang2024robust}. This creates selection pressure toward flexible schedules with recoverable structure, rather than schedules that are only robust when kept fixed. Multi-stage, rolling-horizon, and predictive evaluation extend this idea to time-varying environments, where uncertainty information is updated during execution \cite{wei2024adjustability,bakon2022scheduling}. Prediction is an information-updating mechanism rather than an independent robustness criterion; it affects robust evaluation by updating scenarios, distributions, or uncertainty sets. Stationary uncertainty permits fixed evaluation models, whereas non-stationary uncertainty requires updated or rolling evaluation.

These mechanisms reshape the search behavior of population-based metaheuristics through different selection pressures. Expected-performance evaluation favors average efficiency, worst-case evaluation favors protection under adverse realizations, risk-, regret-, and stability-aware evaluation favors reliability and low sensitivity, distributionally robust evaluation favors resistance to distributional misspecification, and recourse-aware evaluation favors recoverability after uncertainty realization. Robust evaluation therefore determines which regions of the search space are amplified, preserved, or discarded by selection.

\begin{table}[t]
	\centering
	\caption{Robust evaluation, ranking, and selection mechanisms for uncertain BACAP.}
	\label{tab:robust_evaluation_taxonomy}
	\footnotesize
	\renewcommand{\arraystretch}{1.15}
	\begin{tabular}{p{4cm} p{4cm}}
		\toprule
		\textbf{Mechanism} 
		& \textbf{Representative references} \\
		\midrule
		Expected-performance evaluation 
		&  \cite{yan2019dynamic}\cite{ma2021stochastic}\cite{zhen2024primal}\cite{ZHOU2008161}\cite{prayogo2022bi}\cite{li2019integrated}\\
		\midrule
		Worst-case evaluation 
		&  \cite{shang2016robust}\cite{rodrigues2021exact} \cite{CHARGUI2023106251} \cite{liu2020two}\cite{nourmohammadzadeh2022robust} \cite{XIANG2017294}\cite{chargui2023novel}\cite{ran2025two}
		\\
		\midrule
		Risk-, regret-, and stability-aware evaluation 
		& \cite{ji2022enhanced} \cite{rodriguez2014genetic}\cite{kolley2023robust}\cite{xiang2021almost}\cite{prayogo2022bi}
		\cite{park2021particle}\cite{woo2024proactive}\\
		\midrule
		Distributionally robust evaluation 
		&\cite{nie2023distributionally} \cite{AGRA20221} \cite{rodrigues2024handling} \cite{wang2025optimizing}	\cite{wang2024distributionally} \cite{tang2023distributionally}\\
		\midrule
		Recourse-aware and time-adaptive evaluation 
		& \cite{wang2024robust}\cite{zheng2023integrated} \cite{ji2022hybrid}\cite{zheng2024integrated} \cite{IRIS2019365}  \cite{qu2023two} \\
		\bottomrule
	\end{tabular}
\end{table}

\subsection{Robustness-Guided Search Dynamics}
\label{subsec:robustness_guided_search}

The search space of BACAP is highly nonconvex, irregular, and multimodal due to the coupling between berth allocation and QC assignment. After robust evaluation is introduced, the search landscape is no longer determined only by nominal objective values. Instead, it is shaped by uncertainty-dependent performance, worst-case degradation, violation frequency, recoverability, and other robustness-related criteria. As a result, a candidate schedule that appears promising under deterministic evaluation may correspond to a fragile region of the robust search landscape. Population-based metaheuristics must therefore adapt their search dynamics to robustness-aware fitness signals rather than relying only on nominal improvement. In this context, search behavior is no longer aimed solely at improving nominal solution quality, but also at discovering resilient schedule structures, intensifying search around stable or recoverable regions, and dynamically regulating search effort under uncertainty-aware evaluation.

This subsection reviews search dynamics from three complementary perspectives: robustness-guided exploration, robustness-guided exploitation, and robustness-adaptive search regulation. Robustness-guided exploration emphasizes the discovery of structurally diverse and disturbance-resistant scheduling patterns, robustness-guided exploitation focuses on stability-oriented and recoverability-aware local refinement, and robustness-adaptive search regulation balances search effort according to robustness feedback, uncertainty conditions, and search-state information. This organization emphasizes how robust evaluation changes the role of search operators, rather than treating exploration enhancement, diversity maintenance, adaptive control, and hybridization as unrelated algorithmic add-ons.

\subsubsection{Robustness-Guided Exploration}

Robustness-guided exploration expands the search beyond nominally good but fragile schedules and exposes structurally different solutions with distinct uncertainty responses. In BACAP, such exploration can be supported by metaheuristic diversification mechanisms, including Lévy-flight-based search in cuckoo-search variants \cite{aslam2022enhanced}, 
as well as recombination, mutation, neighborhood perturbation, and probabilistic acceptance in other population-based frameworks. These mechanisms diversify vessel service orders, berth allocations, and QC deployment patterns, enabling the population to reach schedule structures that may be less sensitive to arrival deviations, handling-duration fluctuations, or resource disruptions.

Under uncertainty, exploration also needs to cover different response patterns. Two schedules with similar nominal objective values may differ substantially in how they respond to delay propagation, QC failure, or resource congestion. Pareto-based archiving, multi-population search, distributed search, and information-sharing mechanisms help preserve robustness-wise diversity by maintaining schedules that perform well under different uncertainty realizations or robustness criteria \cite{xu2012robust,park2021particle}. Diversity maintenance is therefore treated here as a tool for exploring robust regions, rather than as an isolated algorithmic objective.

A further role of exploration in uncertain BACAP is to identify robust regions rather than isolated robust points. Since robust evaluation aggregates multiple realizations, local neighborhoods may contain schedules with similar average performance but different stability, recoverability, or violation sensitivity. Neighborhood-based robust evaluation and robustness-enhancement mechanisms in evolutionary multiobjective optimization formalize this idea by averaging perturbed neighborhood performance, maintaining robustness-aware elites, and approximating robust Pareto regions rather than only nominal optima \cite{he2019evolutionary,shui2024approximating}. Iterative scenario sampling and statistical testing provide a complementary way to refine robustness estimates and identify stable search regions under uncertainty \cite{starreveld2025robist}.

\subsubsection{Robustness-Guided Exploitation}

Robustness-guided exploitation refines promising schedules using robust fitness information rather than nominal objective values alone. In deterministic metaheuristics, exploitation usually means intensifying the search around solutions with good current objective values. In robust BACAP, local refinement should follow the ranking criterion established in Section~\ref{subsec:robust_evaluation_selection}. If selection is driven by expected performance, exploitation tends to improve average service efficiency across realizations. If it is driven by worst-case evaluation, exploitation tends to reduce extreme losses or critical bottlenecks. If it is driven by risk, regret, or stability criteria, exploitation tends to reduce performance variability or reliability loss.

This type of exploitation is often implemented through local search, neighborhood refinement, or hybrid improvement strategies embedded into metaheuristic search \cite{lau2007robust,shang2016robust}. In BACAP, such mechanisms may refine berth positions, shift service times, adjust QC allocation, or reorder local vessel subsequences to improve robust objective values. The key distinction from deterministic exploitation is that the local search target is a robust fitness landscape \cite{nourmohammadzadeh2024robust}. A move is beneficial not merely because it improves nominal cost, but because it improves aggregated performance, reduces worst-case degradation, or enhances schedule stability \cite{wang2024proactive}.

Robust exploitation may also become critical-realization-oriented. When robust evaluation reveals that a schedule performs poorly under a specific adverse realization, local refinement can focus on that condition. Typical examples include increasing temporal or spatial buffers against delayed arrivals \cite{woo2024proactive}, reducing tight QC coupling in disruption-prone schedules, and modifying local vessel sequences to alleviate conflict under heavy congestion \cite{wang2024proactive}. Such targeted refinement converts robustness evaluation into directed search pressure, consistent with bad-scenario-set, worst-scenario, and learning-scenario neighborhood strategies in robust scheduling, as well as problem-driven scenario generation that concentrates computational effort on risk-relevant realizations \cite{rodrigues2021exact,wang2021two,wang2023bi}.

A closely related form is recoverability-oriented exploitation. In two-stage or recourse-aware settings, a schedule is attractive not only because it has good first-stage performance, but also because it can be repaired or adapted efficiently after uncertainty realization \cite{qu2023two}. Exploitation in this case favors schedules with recoverable structure, such as flexible berth release patterns, moderate QC utilization, or lower repair cost \cite{zheng2024integrated}. This perspective is consistent with recoverable robustness and recovery-to-optimality, where first-stage quality is evaluated jointly with recovery distance or recourse cost, as well as with scenario-guided neighborhood search in robust scheduling that intensifies local improvement toward adverse realizations \cite{bold2025recoverable}.

\subsubsection{Adaptive Search Regulation}

Balancing exploration and exploitation becomes more difficult under uncertainty because robust evaluation is usually more expensive and less stable than deterministic evaluation. Scenario-based or distribution-based fitness estimates may be noisy, worst-case evaluation may push the population too quickly toward conservative regions, and aggressive local refinement may overfit a limited set of realizations. Population-based metaheuristics therefore require adaptive mechanisms that adjust search behavior according to robustness feedback, uncertainty conditions, and search-state information.

Adaptive parameter control is one important mechanism. Existing BACAP studies have adjusted mutation or perturbation intensity, search radius, or operator behavior according to the search stage or current solution quality. In robust search, the same idea can be extended so that exploration is strengthened when the population becomes overly conservative or clustered, and exploitation is strengthened when robustly promising regions have been identified. Adaptive operator selection and adaptive uncertainty sampling can also help stabilize robust search by matching search effort to the uncertainty structure \cite{syberfeldt2010evolutionary,starreveld2025robist}. Such adaptive search is particularly important in robust BACAP because uncertainty evaluation is often computationally expensive, while inappropriate sampling or operator selection may reduce search efficiency or lead to unstable robustness estimation \cite{park2021particle}.

Robustness-adaptive search coordination provides another mechanism for balancing exploration and exploitation under uncertainty. Rather than relying on fixed search behaviors throughout the optimization process, robust metaheuristics may dynamically coordinate global diversification and local refinement according to robustness-aware search feedback. Global population search helps discover structurally diverse berth--QC schedules with different uncertainty responses, whereas embedded local improvement, repair heuristics, or neighborhood refinement can intensify search around robustly promising or recoverable regions \cite{wang2024proactive,liang2025hybrid}. Under uncertainty, such coordination is important because excessive exploration may incur high evaluation cost, while excessive exploitation may prematurely concentrate the search around fragile schedules or overfit a limited set of disturbance realizations.

Structured coordination mechanisms, including hierarchical staged search, decomposition-based optimization, nested optimization, and cooperative search, can further regulate search effort across coupled decision layers. These mechanisms are particularly relevant to BACAP because berth allocation, service timing, and QC assignment are strongly coupled but differ in scale, interaction complexity, and robustness sensitivity \cite{chargui2023novel}. Under uncertainty, disturbances may propagate across subproblems and weaken the effectiveness of isolated optimization. Related studies on cooperative coevolution in noisy environments and nested robust optimization suggest that decomposition quality, interaction learning, and cross-level coordination become increasingly important once uncertainty affects subproblem evaluations and couplings \cite{zhong2023cooperative,wu2022cooperative}.

Overall, robustness-guided search dynamics determine how the population moves over a landscape shaped by uncertainty-aware evaluation. Exploration broadens the coverage of robust schedule structures, exploitation refines schedules according to robust performance, and adaptive regulation balances search effort under expensive and noisy robustness evaluation. These mechanisms jointly influence which robust regions are discovered, intensified, and preserved during the search process.

\begin{table}[t]
	\centering
	\caption{Robustness-guided search dynamics in BACAP metaheuristics.}
	\label{tab:robustness_guided_search}
	\footnotesize
	\renewcommand{\arraystretch}{1.15}
	\begin{tabular}{p{2.5cm} p{5.3cm}}
		\toprule
		\textbf{Search role} 
		& \textbf{Representative references} \\
		\midrule
		Robustness-guided exploration
		& \cite{wu2025ternary}\cite{XIANG2017294}\cite{liang2025hybrid}\cite{alsoufi2016robust}\\
		\midrule
		Robustness-guided exploitation
		&\cite{xu2012robust}\cite{ji2022enhanced}\cite{shang2016robust}\cite{tan2025integrated}\cite{wu2025ternary}\cite{rodriguez2014genetic}\cite{wang2024proactive}\cite{qu2023two}\cite{XIANG2017294}\cite{park2021particle}\cite{nourmohammadzadeh2024robust} \cite{alsoufi2016robust}\cite{dai2023bi} \cite{liu2016decision}\cite{guo2021berth}\cite{rodriguez2014robust}\cite{zhen2015tactical}\\
		\midrule
		Adaptive search regulation
		&\cite{ji2022enhanced}\cite{wang2024proactive}\cite{qu2023two}\cite{XIANG2017294}\cite{park2021particle}\cite{liang2025hybrid}\cite{dai2023bi}\cite{liu2016decision}\cite{guo2021berth}\cite{wu2023integrated}\\
		\bottomrule
	\end{tabular}
\end{table}

\subsection{Feasibility Preservation and Recovery under Uncertainty}
\label{subsec:constraint_handling}

In BACAP, spatial, temporal, and operational constraints define a highly restricted feasible region. Typical constraints include quay-boundary requirements, berth non-overlap, vessel arrival-time restrictions, QC-capacity limits, QC interference, safety distance, and resource exclusivity. These constraints already make deterministic BACAP difficult because small changes in vessel order, berthing position, or QC assignment may lead to infeasible schedules. Under uncertainty, feasibility becomes realization-dependent: a schedule feasible under nominal arrivals and handling durations may become infeasible when vessel arrivals are delayed, handling times fluctuate, or QC availability changes. Constraint handling is therefore not only a feasibility-preservation mechanism, but also a search-control component that influences how the population explores, repairs, and maintains executable schedules under uncertainty.

Constraint-handling mechanisms can be classified according to where they intervene in the metaheuristic search process. Penalty-based methods intervene at the evaluation stage by assigning lower fitness to infeasible solutions. Repair-based methods intervene after solution generation by modifying infeasible individuals. Model- and representation-integrated methods intervene during solution generation by embedding feasibility conditions into the model, representation, decoder, or search operators. Robust feasibility coordination further evaluates whether feasibility can be preserved or recovered across multiple realizations. This intervention-based taxonomy clarifies how each method affects both feasibility preservation and search dynamics.

\subsubsection{Penalty-Based Feasibility Pressure}

Penalty-based methods are widely used because of their simplicity and compatibility with population-based metaheuristics. In this approach, constraint violations are incorporated into the fitness function through penalty terms, thereby discouraging infeasible solutions. Various penalty designs have been applied to BACAP, addressing berth non-overlap, QC conflicts, and other resource constraints \cite{ZHOU2008161}. Some formulations further transform hard- or soft-constraint violations into penalized cost components, including waiting time, delay cost, or operational cost \cite{csahin2016differential}.

From a search perspective, penalty-based handling relaxes the feasible boundary by allowing the population to temporarily explore infeasible regions. This flexibility may improve global exploration, especially in irregular search spaces where strictly feasible solutions are difficult to generate. Under uncertainty, however, penalty design becomes more involved because constraint violations may differ across realizations. A schedule may satisfy berth and QC constraints in most realizations but fail under a small number of severe disruptions. Consequently, the penalty design should specify how violations are aggregated across realizations, for example through average violation, maximum violation, violation frequency, or risk-sensitive violation measures \cite{liang2024evolutionary,chen2020evolutionary}. Different aggregation rules induce different selection pressures: average penalties may tolerate occasional infeasibility, whereas worst-case or risk-sensitive penalties favor schedules that remain feasible under adverse conditions \cite{liu2013efficient,akl2021two}.

\subsubsection{Repair-Based Feasibility Restoration}

Repair-based strategies guide infeasible solutions back into the feasible region through problem-specific correction operators. Some studies combine repair mechanisms with penalty functions and feasibility-oriented heuristics to improve constraint satisfaction~\cite{ji2022enhanced}, while others integrate adaptive repair strategies with the $\varepsilon$-constraint method to progressively tighten feasibility requirements during the search process~\cite{zou2024swarm}. Compared with penalty methods, repair-based approaches directly modify solution structures, which can increase the proportion of feasible individuals and accelerate convergence toward feasible scheduling regions.

In BACAP, repair operations may adjust berth positions, shift service start times, reorder locally conflicting vessels, reassign QCs, or modify QC operating intervals to restore feasibility. Their role becomes particularly important under uncertainty because infeasibility may appear only after a schedule is evaluated under a disturbed realization \cite{li2015real}. Realization-specific repair can bridge robust evaluation and executable scheduling. Once an adverse realization exposes a berth conflict, QC overload, or time-window violation, the repair operator modifies the corresponding part of the schedule \cite{xiang2018reactive,zheng2024integrated}. The main risk is that excessive repair may reduce population diversity or bias the search toward a narrow feasible region. Thus, repair should preserve useful structural variation while correcting operational conflicts.

\subsubsection{Model- and Representation-Integrated Feasibility Preservation}

Model- and representation-integrated handling embeds feasibility conditions into the problem model, solution representation, decoder, or search operators, so that candidate schedules are generated within a restricted feasible or near-feasible region. Rodriguez et al.~\cite{rodriguez2014genetic} incorporated multiple constraint types into genetic representations and operators, reducing the generation of infeasible individuals. More detailed BACAP formulations further introduce tidal windows, tugboat availability, navigation safety, and other practical constraints through auxiliary variables and logical rules.

This mechanism differs from penalty and repair methods. Penalty methods penalize infeasibility after evaluation, whereas repair methods correct infeasibility after solution generation. By contrast, model- and representation-integrated methods prevent or reduce infeasibility during solution generation through feasibility-preserving representations, indirect encodings, and decoder-based schedule construction \cite{londe2025biased,teng2024improved}. For example, a permutation- or path-based representation can ensure that each job or vessel appears exactly once, bounded real or integer variables can keep assignment decisions within admissible ranges, and a decoder can construct schedules sequentially while enforcing resource and precedence constraints. In BACAP, the same logic can be used to maintain berth non-overlap and QC-capacity constraints. Such designs improve feasibility by construction and can reduce post-processing effort. Under uncertainty, however, feasibility embedded under nominal arrivals and handling durations may not guarantee feasibility across disturbed realizations. Integrated handling should therefore be complemented by uncertainty-aware feasibility checking, robust decoding, or recourse-aware repair.

\subsubsection{Robust Feasibility Coordination}

Uncertain BACAP requires distinguishing nominal feasibility, realized feasibility, and robust feasibility. Nominal feasibility means that a schedule satisfies constraints under planned arrivals, planned handling durations, and planned QC availability. Realized feasibility means that the schedule satisfies constraints under a particular operating realization, possibly after repair or recourse. Robust feasibility means that the schedule remains feasible, or can be feasibly recovered, over a prescribed set of realizations. This distinction is important because robust objective performance without robust feasibility is insufficient for practical deployment.

Robust feasibility coordination determines how constraint satisfaction is evaluated and maintained under uncertainty. One approach is feasibility-frequency evaluation, where schedules are favored if they remain feasible in a large fraction of realizations or satisfy prescribed violation probabilities \cite{liu2013efficient,rahman2021scheduling}. Another approach is worst-violation control, where the largest violation or constraint violation under adverse scenarios is explicitly penalized or repaired \cite{wang2021two}. A third approach is recourse-aware feasibility, where a first-stage schedule is accepted if second-stage adjustments can recover feasibility at acceptable cost \cite{van2018combining}. These mechanisms connect constraint handling with robust evaluation and selection: schedules are ranked not only by objective performance, but also by their ability to preserve or recover operational feasibility under uncertainty.

The main trade-off in constraint handling is between feasibility preservation and search flexibility. Penalty-based methods allow broad exploration but require careful penalty calibration. Repair-based methods improve executable-solution quality but may reduce diversity if applied too aggressively. Model- and representation-integrated methods improve feasibility by construction but may restrict the search space and become sensitive to nominal assumptions. Robust feasibility coordination further increases evaluation cost because feasibility must be assessed under multiple realizations or uncertainty descriptions. Effective constraint handling for robust BACAP should integrate feasibility preservation with robustness evaluation and population diversity \cite{rodrigues2021exact}.

\begin{table}[t]
	\centering
	\caption{Constraint-handling mechanisms for robust BACAP.}
	\label{tab:constraint_handling_taxonomy}
	\footnotesize
	\renewcommand{\arraystretch}{1.15}
	\begin{tabular}{p{3.1cm} p{5cm}}
		\toprule
		\textbf{Mechanism} 
		& \textbf{Representative references} \\
		\midrule
		Penalty-based feasibility pressure
		& \cite{JIA2020174}\cite{xiang2021almost}\cite{ZHOU2008161} \cite{XIANG2017294}\cite{liu2016decision}\cite{csahin2016differential}\cite{aslam2023berth}\\
		\midrule
		Repair-based feasibility restoration
		& \cite{zheng2023integrated}\cite{zheng2024integrated} \cite{zou2024swarm}\cite{li2015real} \cite{xiang2018reactive} \cite{ma2015fast}\\
		\midrule
		Model- and representation- integrated preservation
		& \cite{rodriguez2014genetic}\cite{wang2024proactive}\cite{dai2023bi}\cite{liu2016decision}\cite{zhen2012bi}\cite{vacca2013exact}\cite{iris2015integrated}\\
		\midrule
		Robust feasibility coordination
		&  \cite{wang2024robust}\cite{ji2022enhanced}\cite{tan2025integrated}\cite{rodrigues2021exact}\cite{wang2024distributionally}\cite{IRIS2019365}
		\cite{qu2023two}\cite{xiang2021expanded}\\
		\bottomrule
	\end{tabular}
\end{table}

The above review shows that robust population-based metaheuristics for BACAP depend on a sequence of tightly coupled design choices: how candidate schedules are represented and decoded, how uncertain information is converted into evaluation inputs, how robustness changes ranking and selection, how search dynamics respond to robust fitness signals, and how feasibility is preserved or recovered under uncertainty. Since existing studies differ substantially in data assumptions, uncertainty descriptions, robustness criteria, and algorithmic configurations, direct comparison across methods remains difficult. This motivates the benchmark suite introduced in the next section.
\section{Benchmark Suite for BACAP under Uncertainty}
\label{sec:sec_3}

This section presents a benchmark suite for evaluating robust population-based metaheuristics for BACAP under operational uncertainty. The benchmark is designed as a controlled synthetic testbed rather than a reconstruction of a single real terminal. Its purpose is to preserve the main structural elements of BACAP while allowing robustness-oriented metaheuristics to be evaluated under shared and reproducible uncertainty settings.

The benchmark generation procedure follows a two-stage design. In the first stage, a nominal deterministic BACAP instance is generated, including vessel arrival times, vessel lengths, QC capacity, planning horizon, and tidal windows. This nominal instance defines the underlying scheduling structure and is consistent with the deterministic BACAP formulation introduced in Section~\ref{sec:basic_structure}. In the second stage, uncertain realizations are generated from the nominal instance by perturbing vessel arrival times, handling durations, and QC availability. Each realization represents one possible operational environment under which a candidate schedule is evaluated. For a given benchmark case, the same set of uncertain realizations is shared by all algorithms to support controlled comparison.

\subsection{Data Generator}

To emulate the operational variability observed in container terminals, the proposed generator considers three major sources of uncertainty: vessel-arrival deviations, handling-duration fluctuations, and QC availability~\cite{bierwirth2010survey}. These uncertainties are associated with maritime traffic conditions, weather states, equipment disruptions, and time-varying operational efficiency. Tidal windows are also incorporated as service-start feasibility constraints, so that vessel operations can begin only within admissible tidal intervals~\cite{du2015modeling}.

\subsubsection{Uncertainty in Vessel Arrival Time}

The actual arrival time of vessel \(i\) is influenced by maritime traffic, weather conditions, and vessel-specific performance. Following stochastic and robust berth-planning studies that model uncertain vessel arrivals~\cite{yan2019dynamic,SCHEPLER2019363,umang2017real}, the actual arrival time is generated as
\begin{equation}
	\label{eq:arrival_uncertainty}
	t_i = a_i + f_w(\eta_w(a_i)) + M_1\xi_i + M_2\zeta_i .
\end{equation}
Here, \(a_i\) is the nominal arrival time, \(f_w(\eta_w(a_i))\) is the weather-induced arrival delay evaluated at the nominal arrival time, \(\xi_i\sim\operatorname{Pois}(\lambda)\) models the number of congestion events during the vessel voyage, and \(\zeta_i\sim\mathcal{N}_{+}(\theta_v,\varrho_v^2)\) denotes a non-negative truncated normal disturbance for vessel-specific performance variation. The coefficients \(M_1\) and \(M_2\) scale the traffic-related and vessel-specific delay terms.

The weather state \(\eta_w(t)\in\{0,1,2\}\) corresponds to normal weather, slightly adverse weather, and severely adverse weather. Following the general Markov-chain-based weather generation idea in~\cite{yang2011first}, the weather state evolves according to
\[
P_{\mathrm{w}}=
\begin{bmatrix}
	p^{w}_{00} & p^{w}_{01} & p^{w}_{02}\\
	p^{w}_{10} & p^{w}_{11} & p^{w}_{12}\\
	p^{w}_{20} & p^{w}_{21} & p^{w}_{22}
\end{bmatrix},
\]
where \(p^{w}_{rs}=\mathbb{P}(\eta_w(t+1)=s\mid \eta_w(t)=r)\), and each row sums to one. The weather-delay mapping is defined by \(f_w(\eta_w(t))=\delta^{w}_{r}\) when \(\eta_w(t)=r\), where \(r\in\{0,1,2\}\) and \(\delta^{w}_{r}\in\mathbb{R}_{+}\) is the state-dependent arrival-delay value.

\subsubsection{Uncertainty of Handling Durations}

Vessel handling durations are affected by cargo characteristics, weather conditions, and operational efficiency~\cite{guo2021berth,HE2021101252,TANG2024110038}. Let \(h_i^0=d_i-s_i\) denote the nominal handling duration of vessel \(i\). The realized handling duration is modeled as
\begin{equation}
	\label{eq:handling_uncertainty}
	\tilde h_i = h_i^0 + g_w(\eta_w(s_i)) + g_e(\eta_e(s_i)) + M_3\vartheta_i .
\end{equation}
Here, \(g_w(\eta_w(s_i))\) is the weather-induced handling delay, \(g_e(\eta_e(s_i))\) is the operational-efficiency-induced handling delay, and \(\vartheta_i\sim\mathcal{N}_{+}(\theta_c,\varrho_c^2)\) denotes a non-negative stochastic handling disturbance. The coefficient \(M_3\) scales the stochastic handling disturbance.

The operational-efficiency state \(\eta_e(t)\in\{0,1,2\}\) corresponds to high, moderate, and low productivity levels. Its evolution is modeled by a Markov transition matrix following Markov-based terminal-efficiency modeling~\cite{mennis2008improving}:
\[
P_{\mathrm{e}}=
\begin{bmatrix}
	p^{e}_{00} & p^{e}_{01} & p^{e}_{02}\\
	p^{e}_{10} & p^{e}_{11} & p^{e}_{12}\\
	p^{e}_{20} & p^{e}_{21} & p^{e}_{22}
\end{bmatrix},
\]
where \(p^{e}_{rs}=\mathbb{P}(\eta_e(t+1)=s\mid \eta_e(t)=r)\), and each row sums to one. The weather-induced handling delay and the efficiency-induced handling delay are defined by \(g_w(\eta_w(t))=\delta^{h,w}_{r}\) when \(\eta_w(t)=r\), and \(g_e(\eta_e(t))=\delta^{e}_{r}\) when \(\eta_e(t)=r\), where \(r\in\{0,1,2\}\). The values \(\delta^{h,w}_{r}\) and \(\delta^{e}_{r}\) are non-negative state-dependent delay values.

\subsubsection{Uncertainty of QC Availability}

QC availability is represented by a binary state \(A_{k,t}\in\{0,1\}\), where \(A_{k,t}=0\) indicates that QC \(k\) is unavailable  at time $t$ due to maintenance or failure, and \(A_{k,t}=1\) indicates normal operation~\cite{li2019integrated,krimi2020modelling}. To emulate stochastic switching between these two states, a two-state Markov chain is used, consistent with crane reliability modeling in~\cite{mohammadi2021risk}:
\[
P_{\mathrm{q}}=
\begin{bmatrix}
	p^{q}_{00} & p^{q}_{01}\\
	p^{q}_{10} & p^{q}_{11}
\end{bmatrix},
\]
where \(p^{q}_{rs}\) denotes the probability of moving from state \(r\) to state \(s\) in the next period. If \(\mathcal{Q}^{\mathrm{plan}}_i\) denotes the set of QCs planned for vessel \(i\), the realized number of operational QCs during service is obtained from the availability states of QCs in \(\mathcal{Q}^{\mathrm{plan}}_i\). This quantity is denoted by \(q_i^{\mathrm{real}}\) and is used in the evaluation metrics.

\subsubsection{Tidal Windows}

Tidal restrictions are represented by a set of admissible service-start windows~\cite{jiao2018integrated},
\[
\mathcal{W}=\{[\underline{t}_g,\overline{t}_g]\}_{g=1}^{G}.
\]
Each interval \([\underline{t}_g,\overline{t}_g]\) denotes a valid time window during which vessel service may start. A feasible service start satisfies \(s_i\in\bigcup_{g=1}^{G}[\underline{t}_g,\overline{t}_g]\). If a planned service start falls outside all admissible intervals, it is postponed to the start of the next available window.

\subsection{Performance Evaluation Metrics}

The benchmark uses two metrics to evaluate scheduling efficiency and robustness. Average port time measures the operational cost of a schedule under realized uncertainty, while survival time evaluates the schedule's slack-based robustness. Detailed computational procedures, including the evaluation of the adjusted service duration, tidal waiting time, and survival-time components, are provided in Section~S-I of the supplementary document.

\subsubsection{Average Port Time}

The port-time indicator combines QC-availability-adjusted service duration, tidal waiting time, and arrival-related service mismatch. Let \(D_i\) denote the adjusted service duration of vessel \(i\), \(W_i^{\mathrm{tide}}\) denote its tidal waiting time, and \(W_i^{\mathrm{arr}}=|a_i-s_i|\) denote its arrival-related service mismatch. The evaluated port time of vessel \(i\) is
\begin{equation}
	\label{eq:port_time}
	T_i^{\mathrm{port}} = D_i + W_i^{\mathrm{tide}} + W_i^{\mathrm{arr}} .
\end{equation}
The adjusted service duration \(D_i\) is computed according to the realized number of operational QCs:
\begin{equation}
	\label{eq:adjusted_service_duration}
	D_i =
	\begin{cases}
		(d_i-s_i)\dfrac{q_i^{\mathrm{plan}}}{q_i^{\mathrm{real}}}, & q_i^{\mathrm{real}}>0,\\[0.8em]
		\kappa(d_i-s_i), & q_i^{\mathrm{real}}=0,
	\end{cases}
\end{equation}
where \(q_i^{\mathrm{plan}}\) is the planned number of QCs assigned to vessel \(i\), \(q_i^{\mathrm{real}}\) is the realized number of operational QCs, and \(\kappa>1\) is a penalty coefficient for complete QC unavailability. The average port time is
\begin{equation}
	\label{eq:avg_port_time}
	\bar{T}_{\mathrm{port}}=\frac{1}{|\mathcal{V}|}\sum_{i\in\mathcal{V}}T_i^{\mathrm{port}} .
\end{equation}
Smaller values of \(\bar{T}_{\mathrm{port}}\) indicate better operational efficiency.

\subsubsection{Survival Time}

Survival time measures a schedule's ability to absorb disturbances before operational conflicts occur. This interpretation is related to robustness definitions in dynamic optimization, where survival time describes how long a solution remains feasible or maintains acceptable performance under changing conditions~\cite{fu2014robust}. In the BACAP setting, survival time is used to quantify the slack of a berth-crane schedule against tidal restrictions, berth-space interference, and crane-sharing conflicts~\cite{bierwirth2010survey,himmiche2023robustness}.

For vessel \(i\), the survival time \(\sigma_i\) is defined as an additive slack-based measure consisting of the tidal buffer, the berth-related temporal margin, and the crane-related temporal margin. The tidal buffer measures the remaining time before the current tidal window closes. The berth-related margin measures temporal separation from vessels occupying overlapping berth segments. The crane-related margin measures temporal separation from vessels sharing at least one assigned QC. Larger positive margins indicate greater slack, while negative margins indicate insufficient temporal separation. Detailed component definitions are provided in Section~S-I of the supplementary document.

The survival threshold is defined as the empirical median of nominal service durations:
\[
\theta=P_{50}\left(\{d_i-s_i:i\in\mathcal{V}\}\right).
\]
The set of robustness-effective vessels is \(S=\{i\in\mathcal{V}:\sigma_i>\theta\}\). The robustness index is
\begin{equation}
	\label{eq:robustness_index}
	\bar{\sigma}=
	\begin{cases}
		\displaystyle \frac{1}{|S|}\sum_{i\in S}\sigma_i, & S\neq\emptyset,\\[1.0em]
		0, & S=\emptyset.
	\end{cases}
\end{equation}
This index focuses on vessels whose survival time exceeds a representative service-duration threshold. A larger \(\bar{\sigma}\) indicates stronger slack-based robustness.

\subsection{Benchmark Cases}

The benchmark suite includes 16 cases obtained by combining four vessel-arrival patterns with four uncertainty scenarios, as shown in Table~\ref{tab:benchmark_cases}. Three problem scales, 50 vessels, 100 vessels, and 200 vessels, are considered separately for each case. The vessel-arrival patterns include four synthetic fluctuation patterns adapted from stochastic berth-planning studies and dynamic-optimization benchmark design~\cite{fu2014robust}.

\textit{Uniform pattern.} The actual arrival time is generated as
\begin{equation}
	\label{eq:uniform_arrival}
	t_i=a_i+\delta_i,\qquad \delta_i\sim\mathcal{U}(-\Delta_{\max},\Delta_{\max}),
\end{equation}
where \(\delta_i\) is the arrival-time deviation and \(\Delta_{\max}\) is the maximum absolute deviation from the nominal arrival time.

\textit{Gaussian pattern.} The actual arrival time is generated as
\begin{equation}
	\label{eq:gaussian_arrival}
	t_i=a_i+\epsilon_i,\qquad \epsilon_i\sim\mathcal{N}(0,\sigma^2),
\end{equation}
where \(\epsilon_i\) is a normally distributed arrival-time perturbation.

\textit{Chaotic-inspired nonlinear pattern.} The actual arrival time is generated as
\begin{equation}
	\label{eq:chaotic_arrival}
	t_i=a^{\min}+A(u_i-a^{\min})
	\left(1-\frac{u_i-a^{\min}}{a^{\max}-a^{\min}}\right),
\end{equation}
where \(u_i\sim\mathcal{U}(a^{\min},a^{\max})\), \(a^{\min}=\min_{i\in\mathcal{V}}a_i-2\), \(a^{\max}=\max_{i\in\mathcal{V}}a_i+2\), and \(A=3.67\).

\textit{Periodic pattern.} The actual arrival time is generated as
\begin{equation}
	\label{eq:periodic_arrival}
	t_i=a^{\min}+\frac{a^{\max}-a^{\min}}{2}
	\left[\sin\left(\frac{2\pi}{P}i+\phi\right)+1\right],
\end{equation}
where \(P\) is the period length and \(\phi=\pi/4\) is the phase offset.

The uncertainty scenarios represent different sources of operational variability. US1 perturbs vessel arrivals only, US2 perturbs handling durations only, US3 perturbs QC availability only, and US4 combines vessel-arrival, handling-duration, and QC-availability uncertainties.

\begin{table}[htbp]
	\caption{Benchmark cases constructed from four arrival patterns and four uncertainty scenarios.}
	\centering
	\footnotesize
	\renewcommand{\arraystretch}{1.1}
	\begin{tabular}{clc clc}
		\toprule
		\multicolumn{3}{c}{\textbf{Cases C1--C8}} & \multicolumn{3}{c}{\textbf{Cases C9--C16}} \\
		\cmidrule(r){1-3} \cmidrule(l){4-6}
		\textbf{ID} & \textbf{Pattern} & \textbf{US} & \textbf{ID} & \textbf{Pattern} & \textbf{US} \\
		\midrule
		C1  & Uniform   & \multirow{4}{*}{US1} & C9  & Uniform   & \multirow{4}{*}{US3} \\
		C2  & Gaussian  &                      & C10 & Gaussian  &                      \\
		C3  & Chaotic   &                      & C11 & Chaotic   &                      \\
		C4  & Periodic  &                      & C12 & Periodic  &                      \\
		\cmidrule(r){1-3} \cmidrule(l){4-6}
		C5  & Uniform   & \multirow{4}{*}{US2} & C13 & Uniform   & \multirow{4}{*}{US4} \\
		C6  & Gaussian  &                      & C14 & Gaussian  &                      \\
		C7  & Chaotic   &                      & C15 & Chaotic   &                      \\
		C8  & Periodic  &                      & C16 & Periodic  &                      \\
		\bottomrule
	\end{tabular}
	\label{tab:benchmark_cases}
\end{table}

To support reproducibility, the related MATLAB source code and supplementary materials are provided in an online repository: \url{https://github.com/EvoNexusX/2026LiBACAPSURVSY}.

\section{Baseline Evaluation on the Proposed Benchmark}
\label{sec:sec_4}

This section reports illustrative baseline experiments on the proposed BACAP benchmark suite. Three representative population-based metaheuristics are considered: GA, ACO, and PSO. Each metaheuristic is combined with three robustness strategies, namely min--max robust optimization (MMRO), expected robust optimization (ERO), and two-stage robust optimization (TSRO), resulting in nine algorithm--strategy combinations. To keep the comparison controlled, all algorithms use a population, colony, or swarm size of 50 and run for 100 generations or iterations. For GA, the crossover probability is set to 0.9, the mutation probability is set to 0.2, and roulette-wheel selection with elitism is used. For ACO, the archive size is set to 50, and the initial perturbation strength is set to 0.2 with exponential decay. For PSO, the inertia weight is set to \(w=0.7\), and the cognitive and social coefficients are set to \(c_1=c_2=1.5\). The same benchmark instances and uncertainty realizations are used for all algorithms within each benchmark case.

Each algorithm--strategy combination is independently executed 25 times on the 16 benchmark cases under each problem scale. The results are reported in terms of average port time \(\bar{T}_{\mathrm{port}}\) and survival-time robustness \(\bar{\sigma}\), which measure scheduling efficiency and robustness, respectively. For each metric, the mean and standard deviation are reported to summarize performance and run-to-run variability under uncertain operational conditions.

\begin{table*}[htbp]
	\centering
	\scriptsize
	\caption{Baseline results (mean $\pm$ std) of average port time and survival-time robustness for nine robust population-based metaheuristics under the 100-vessel setting.}
	\label{tab:100_A}
	\resizebox{\textwidth}{!}{
		\begin{tabular}{lllccccccccc}
			\toprule
			Cases & Metric &Value& GA-MMRO & GA-ERO & GA-TSRO & ACO-MMRO & ACO-ERO & ACO-TSRO & PSO-MMRO & PSO-ERO & PSO-TSRO \\
			\midrule
			
			C1 & $\bar{T}_{\mathrm{port}}$&Mean & 3.14E+01 & 3.08E+01 & 3.08E+01 & 2.90E+01 & 2.85E+01 & 2.88E+01 & 2.83E+01 & 2.76E+01 & \textbf{2.71E+01} \\
			&              & Std&1.35E+00 & 7.00E-01 & 7.90E-01 & 9.80E-01 & 1.04E+00 & 9.90E-01 & 1.36E+00 & 1.03E+00 & \textbf{1.02E+00} \\
			& $\bar{\sigma}$ & Mean &7.10E+00 & 7.02E+00 & 7.02E+00 & 7.14E+00 & 7.01E+00 & 7.03E+00 & 7.13E+00 & \textbf{7.22E+00} & 7.05E+00 \\
			&              & Std&3.70E-01 & 3.50E-01 & 3.50E-01 & 3.40E-01 & 2.90E-01 & 2.60E-01 & 3.20E-01 & \textbf{4.40E-01} & 3.80E-01 \\
			C2 & $\bar{T}_{\mathrm{port}}$ &Mean & 3.19E+01 & 3.09E+01 & 3.12E+01 & 2.92E+01 & 2.87E+01 & 2.85E+01 & 2.86E+01 & \textbf{2.74E+01} & 2.82E+01 \\
			&              & Std&1.21E+00 & 7.90E-01 & 9.10E-01 & 7.10E-01 & 9.00E-01 & 8.50E-01 & 1.39E+00 & \textbf{9.50E-01} & 8.20E-01 \\
			& $\bar{\sigma}$ & Mean &7.09E+00 & 6.97E+00 & 6.95E+00 & 7.03E+00 & 7.05E+00 & 7.10E+00 & 6.91E+00 & \textbf{7.16E+00} & 6.97E+00 \\
			&              & Std&3.80E-01 & 4.10E-01 & 3.70E-01 & 3.30E-01 & 3.30E-01 & 4.30E-01 & 4.00E-01 & \textbf{3.90E-01} & 3.20E-01 \\
			C3 & $\bar{T}_{\mathrm{port}}$ & Mean &3.57E+01 & 3.53E+01 & 3.55E+01 & 3.42E+01 & 3.36E+01 & 3.33E+01 & 3.30E+01 & \textbf{3.19E+01} & 3.21E+01 \\
			&              & Std&7.90E-01 & 7.60E-01 & 7.60E-01 & 8.40E-01 & 7.30E-01 & 6.30E-01 & 1.60E+00 & \textbf{1.23E+00} & 1.11E+00 \\
			& $\bar{\sigma}$ & Mean &6.95E+00 & 7.05E+00 & 7.09E+00 & 6.93E+00 & 6.93E+00 & 7.00E+00 & \textbf{7.06E+00} & 6.96E+00 & 6.93E+00 \\
			&              & Std&3.80E-01 & 4.40E-01 & 3.80E-01 & 2.90E-01 & 3.00E-01 & 3.50E-01 & \textbf{3.70E-01} & 4.10E-01 & 3.50E-01 \\
			C4 & $\bar{T}_{\mathrm{port}}$ & Mean &3.59E+01 & 3.54E+01 & 3.50E+01 & 3.39E+01 & 3.35E+01 & 3.34E+01 & 3.38E+01 & 3.32E+01 & \textbf{3.29E+01} \\
			&              & Std&1.11E+00 & 8.90E-01 & 1.10E+00 & 7.30E-01 & 6.80E-01 & 8.40E-01 & 1.03E+00 & 7.80E-01 & \textbf{8.60E-01} \\
			& $\bar{\sigma}$ & Mean &6.93E+00 & 7.01E+00 & 7.11E+00 & 7.08E+00 & 7.01E+00 & \textbf{7.14E+00} & 6.77E+00 & 6.93E+00 & 6.96E+00 \\
			&              & Std&3.30E-01 & 3.70E-01 & 3.00E-01 & 3.30E-01 & 3.90E-01 & \textbf{2.80E-01} & 4.00E-01 & 4.60E-01 & 3.30E-01 \\
			\midrule
			
			C5 & $\bar{T}_{\mathrm{port}}$ & Mean &3.32E+01 & 3.24E+01 & 3.23E+01 & 3.06E+01 & 2.97E+01 & 2.99E+01 & 3.00E+01 & 2.86E+01 & \textbf{2.85E+01} \\
			&              & Std&8.80E-01 & 1.00E+00 & 1.10E+00 & 9.20E-01 & 8.80E-01 & 8.80E-01 & 1.57E+00 & 1.01E+00 & \textbf{1.18E+00} \\
			& $\bar{\sigma}$ & Mean &7.34E+00 & 7.37E+00 & 7.38E+00 & 7.49E+00 & 7.48E+00 & 7.45E+00 & 7.50E+00 & 7.52E+00 & \textbf{7.54E+00} \\
			&              & Std&4.40E-01 & 3.70E-01 & 3.20E-01 & 3.20E-01 & 3.80E-01 & 2.50E-01 & 3.10E-01 & 3.20E-01 & \textbf{3.40E-01} \\
			C6 & $\bar{T}_{\mathrm{port}}$ & Mean &3.30E+01 & 3.19E+01 & 3.21E+01 & 3.06E+01 & 3.01E+01 & 3.01E+01 & 2.98E+01 & \textbf{2.84E+01} & 2.88E+01 \\
			&              & Std&1.08E+00 & 8.00E-01 & 7.30E-01 & 9.10E-01 & 8.70E-01 & 9.70E-01 & 1.13E+00 & \textbf{1.13E+00 }& 8.70E-01 \\
			& $\bar{\sigma}$ & Mean &7.37E+00 & 7.39E+00 & 7.44E+00 & 7.43E+00 & \textbf{7.60E+00} & 7.47E+00 & 7.43E+00 & 7.39E+00 & 7.44E+00 \\
			&              & Std&4.20E-01 & 3.30E-01 & 3.90E-01 & 2.30E-01 & \textbf{3.50E-01} & 3.70E-01 & 3.90E-01 & 2.90E-01 & 3.40E-01 \\
			C7 & $\bar{T}_{\mathrm{port}}$ & Mean &3.49E+01 & 3.44E+01 & 3.41E+01 & 3.35E+01 & 3.27E+01 & 3.24E+01 & 3.36E+01 & \textbf{3.15E+01} & 3.15E+01 \\
			&              & Std&8.00E-01 & 1.05E+00 & 9.00E-01 & 1.03E+00 & 6.90E-01 & 7.60E-01 & 1.10E+00 & \textbf{8.40E-01} & 9.80E-01 \\
			& $\bar{\sigma}$ & Mean &7.30E+00 & 7.30E+00 & 7.29E+00 & \textbf{7.45E+00} & 7.24E+00 & 7.35E+00 & 7.28E+00 & 7.28E+00 & 7.26E+00 \\
			&              & Std&4.40E-01 & 3.80E-01 & 4.00E-01 & \textbf{3.90E-01} & 3.70E-01 & 4.60E-01 & 4.50E-01 & 3.50E-01 & 4.80E-01 \\
			C8 & $\bar{T}_{\mathrm{port}}$ & Mean &3.22E+01 & 3.13E+01 & 3.16E+01 & 3.11E+01 & 3.06E+01 & 3.05E+01 & 3.16E+01 & \textbf{3.03E+01} & 3.04E+01 \\
			&              & Std&9.40E-01 & 8.30E-01 & 8.00E-01 & 6.60E-01 & 9.10E-01 & 6.80E-01 & 9.40E-01 & \textbf{7.20E-01} & 8.70E-01 \\
			& $\bar{\sigma}$ & Mean &7.35E+00 & 7.42E+00 & 7.40E+00 & 7.43E+00 & 7.43E+00 & \textbf{7.54E+00} & 7.36E+00 & 7.35E+00 & 7.41E+00 \\
			&              & Std&4.30E-01 & 4.30E-01 & 3.00E-01 & 3.40E-01 & 3.70E-01 & \textbf{3.40E-01} & 3.30E-01 & 4.30E-01 & 2.70E-01 \\
			\midrule
			
			C9 & $\bar{T}_{\mathrm{port}}$ & Mean &3.50E+01 & 3.40E+01 & 3.40E+01 & 3.22E+01 & 3.14E+01 & 3.16E+01 & 3.17E+01 & 3.03E+01 & \textbf{3.02E+01} \\
			&              & Std&9.10E-01 & 9.30E-01 & 1.14E+00 & 9.30E-01 & 9.00E-01 & 9.10E-01 & 1.56E+00 & 1.08E+00 & \textbf{1.23E+00} \\
			& $\bar{\sigma}$ & Mean &6.70E+00 & 6.72E+00 & 6.70E+00 & 6.86E+00 & 6.83E+00 & \textbf{6.86E+00} & 6.67E+00 & 6.75E+00 & 6.80E+00 \\
			&              & Std&3.70E-01 & 4.10E-01 & 2.30E-01 & 3.10E-01 & 4.40E-01 & \textbf{4.50E-01} & 3.60E-01 & 4.30E-01 & 2.70E-01 \\
			C10 & $\bar{T}_{\mathrm{port}}$ & Mean &3.48E+01 & 3.36E+01 & 3.38E+01 & 3.22E+01 & 3.18E+01 & 3.19E+01 & 3.15E+01 & \textbf{3.01E+01} & 3.05E+01 \\
			&              & Std&1.04E+00 & 8.30E-01 & 7.20E-01 & 9.80E-01 & 8.30E-01 & 9.50E-01 & 1.25E+00 & \textbf{1.15E+00} & 9.00E-01 \\
			& $\bar{\sigma}$ & Mean &6.72E+00 & 6.78E+00 & 6.77E+00 & 6.78E+00 & 6.78E+00 & \textbf{6.86E+00} & 6.65E+00 & 6.84E+00 & 6.70E+00 \\
			&              & Std&3.70E-01 & 3.70E-01 & 3.70E-01 & 3.70E-01 & 3.10E-01 & \textbf{3.40E-01} & 3.40E-01 & 3.90E-01 & 3.90E-01 \\
			C11 & $\bar{T}_{\mathrm{port}}$ & Mean &3.66E+01 & 3.61E+01 & 3.58E+01 & 3.52E+01 & 3.44E+01 & 3.41E+01 & 3.54E+01 & \textbf{3.31E+01} & 3.31E+01 \\
			&              & Std&7.90E-01 & 9.30E-01 & 8.00E-01 & 1.09E+00 & 6.50E-01 & 7.90E-01 & 1.10E+00 & \textbf{8.90E-01} & 9.90E-01 \\
			& $\bar{\sigma}$ & Mean &6.62E+00 & 6.73E+00 & 6.80E+00 & \textbf{6.81E+00} & 6.76E+00 & 6.83E+00 & 6.73E+00 & 6.72E+00 & 6.55E+00 \\
			&              & Std&4.10E-01 & 3.80E-01 & 4.40E-01 & \textbf{3.60E-01} & 4.90E-01 & 3.90E-01 & 3.90E-01 & 3.30E-01 & 3.20E-01 \\
			C12 & $\bar{T}_{\mathrm{port}}$ & Mean &3.39E+01 & 3.30E+01 & 3.33E+01 & 3.28E+01 & 3.24E+01 & 3.22E+01 & 3.32E+01 & \textbf{3.21E+01} & 3.21E+01 \\
			&              & Std&9.30E-01 & 8.50E-01 & 7.40E-01 & 6.70E-01 & 9.00E-01 & 7.40E-01 & 1.01E+00 & \textbf{6.90E-01} & 8.60E-01 \\
			& $\bar{\sigma}$ & Mean &6.72E+00 & 6.72E+00 & \textbf{6.79E+00} & 6.74E+00 & 6.78E+00 & 6.77E+00 & 6.73E+00 & 6.74E+00 & 6.62E+00 \\
			&              & Std&3.70E-01 & 3.10E-01 & \textbf{4.60E-01} & 3.70E-01 & 3.00E-01 & 2.50E-01 & 3.70E-01 & 4.00E-01 & 3.10E-01 \\
			\midrule
			
			C13 & $\bar{T}_{\mathrm{port}}$ & Mean &3.45E+01 & 3.47E+01 & 3.45E+01 & 3.37E+01 & 3.10E+01 & 3.11E+01 & 3.42E+01 & 3.00E+01 & \textbf{2.99E+01} \\
			&              & Std&7.50E-01 & 7.40E-01 & 6.90E-01 & 7.50E-01 & 9.00E-01 & 9.10E-01 & 9.10E-01 & 1.02E+00 & \textbf{1.23E+00} \\
			& $\bar{\sigma}$ & Mean &7.04E+00 & 6.95E+00 & 7.04E+00 & 7.40E+00 & 7.39E+00 & 7.43E+00 & 7.27E+00 & 7.36E+00 & \textbf{7.57E+00} \\
			&              & Std&3.00E-01 & 3.30E-01 & 4.30E-01 & 3.90E-01 & 4.60E-01 & 3.90E-01 & 4.20E-01 & 3.70E-01 & \textbf{3.20E-01} \\
			C14 & $\bar{T}_{\mathrm{port}}$ & Mean &3.47E+01 & 3.44E+01 & 3.48E+01 & 3.40E+01 & 3.14E+01 & 3.14E+01 & 3.44E+01 & \textbf{2.98E+01} & 3.02E+01 \\
			&              & Std&8.50E-01 & 1.04E+00 & 7.20E-01 & 9.50E-01 & 8.30E-01 & 9.50E-01 & 9.80E-01 & \textbf{1.12E+00} & 8.60E-01 \\
			& $\bar{\sigma}$ & Mean &6.97E+00 & 6.91E+00 & 6.99E+00 & 7.37E+00 & 7.41E+00 & \textbf{7.54E+00} & 7.37E+00 & 7.42E+00 & 7.35E+00 \\
			&              & Std&3.70E-01 & 4.90E-01 & 4.50E-01 & 3.90E-01 & 4.30E-01 & \textbf{3.70E-01} & 3.80E-01 & 4.00E-01 & 3.90E-01 \\
			C15 & $\bar{T}_{\mathrm{port}}$ & Mean &3.67E+01 & 3.66E+01 & 3.64E+01 & 3.63E+01 & 3.41E+01 & 3.38E+01 & 3.64E+01 & \textbf{3.29E+01} & 3.29E+01 \\
			&              & Std&1.00E+00 & 8.70E-01 & 9.80E-01 & 1.07E+00 & 6.60E-01 & 8.10E-01 & 7.40E-01 & \textbf{8.80E-01} & 9.90E-01 \\
			& $\bar{\sigma}$ & Mean &7.11E+00 & 6.87E+00 & 6.98E+00 & 7.38E+00 & 7.30E+00 & \textbf{7.47E+00} & 7.37E+00 & 7.41E+00 & 7.33E+00 \\
			&              & Std&3.50E-01 & 4.60E-01 & 4.10E-01 & 4.70E-01 & 4.00E-01 & \textbf{3.20E-01} & 4.10E-01 & 3.80E-01 & 3.30E-01 \\
			C16 & $\bar{T}_{\mathrm{port}}$ & Mean &3.39E+01 & 3.41E+01 & 3.41E+01 & 3.42E+01 & 3.22E+01 & 3.21E+01 & 3.34E+01 & \textbf{3.21E+01} & 3.21E+01 \\
			&              & Std&7.40E-01 & 7.80E-01 & 9.10E-01 & 1.02E+00 & 9.00E-01 & 7.20E-01 & 1.01E+00 & \textbf{6.60E-01} & 8.30E-01 \\
			& $\bar{\sigma}$ & Mean &7.25E+00 & 7.05E+00 & 7.11E+00 & 7.55E+00 & 7.45E+00 & \textbf{7.54E+00} & 7.35E+00 & 7.37E+00 & 7.36E+00 \\
			&              & Std&4.90E-01 & 3.70E-01 & 4.10E-01 & 4.80E-01 & 4.10E-01 & \textbf{3.80E-01} & 3.40E-01 & 4.00E-01 & 4.30E-01 \\
			\bottomrule
		\end{tabular}
	}
\end{table*}

Table~\ref{tab:100_A} reports the performance of the nine robust population-based metaheuristics under the 100-vessel setting. The results for the 50-vessel and 200-vessel settings are provided in the supplementary material. A lower \(\bar{T}_{\mathrm{port}}\) indicates higher scheduling efficiency, whereas a higher \(\bar{\sigma}\) indicates stronger robustness. For the mean rows, the best result is highlighted in bold, while standard deviations are reported to indicate performance variability.

The illustrative baseline results show that PSO-based combinations often achieve lower average port times than GA- and ACO-based combinations in the 100-vessel setting. In particular, PSO-ERO and PSO-TSRO obtain competitive efficiency across several uncertainty scenarios. In terms of robustness, ACO-based combinations remain competitive in some cases, especially when combined with MMRO or TSRO. These observations suggest that the interaction between the search mechanism and the robustness strategy substantially affects the efficiency--robustness trade-off. The results therefore provide reference baselines for the proposed benchmark rather than a definitive ranking of metaheuristic algorithms.

\section{Challenges, Future Research, and Conclusion}
\label{sec:sec_5}

\subsection{Challenges and Future Research}

Although robust population-based metaheuristics have shown promise for uncertain BACAP, several challenges remain in benchmark extension, robustness-aware search design, dynamic robustness maintenance, and non-stationary uncertainty handling.

\subsubsection{Benchmark Extension and Real-World Validation}

This paper takes a step toward reproducible evaluation by providing a controlled benchmark suite, generation details, and baseline results for uncertain BACAP. Nevertheless, a benchmark suite is not a fixed endpoint. Future research can extend the benchmark toward richer operational settings, including heterogeneous terminal layouts, multiple quay areas, vessel-class-dependent handling requirements, detailed QC interference rules, and more realistic tidal or navigation constraints. Another important direction is to calibrate benchmark parameters using anonymized real terminal data, so that synthetic instances remain structurally controlled while better reflecting practical operating conditions. Community-level validation is also needed to establish common evaluation protocols, including fixed random seeds, shared uncertainty realizations, runtime reporting, statistical tests, and convergence profiles. Such extensions would help transform benchmark-based evaluation from a paper-specific comparison into a more stable basis for assessing robust metaheuristic designs.

\subsubsection{Robustness-Aware Metaheuristic Search}

Many existing studies incorporate robustness mainly through evaluation or selection, while the search operators themselves often remain close to their deterministic counterparts. Conventional crossover, mutation, and repair operators may not explicitly account for the sensitivity of scheduling structures to uncertainty propagation and temporal coupling. As a result, offspring may lose useful robustness-related structures, such as stable berthing patterns, temporal buffers, or recoverable QC assignments~\cite{jin2005evolutionary}.

Future research can develop robustness-aware metaheuristic search frameworks that embed robustness information into variation, selection, and repair. Historical robustness experience may guide populations toward robust regions by identifying structures that resist delays and conflicts~\cite{du2018high,meng2021hip}. Promising directions include memory-guided operators that preserve stable substructures, adaptive operator selection based on robustness feedback~\cite{li2013adaptive}, and learning-assisted mechanisms, such as reinforcement learning or pattern mining, for discovering robust structural regularities in large-scale stochastic scheduling spaces~\cite{yang2024reinforcement}.

\subsubsection{Robust Optimization Over Time for BACAP}

Robustness under time-evolving uncertainty requires scheduling methods to account for the cumulative effects of vessel delays, weather shifts, workload fluctuations, and QC disruptions. A temporally robust schedule should preserve feasibility and service quality across the planning horizon while limiting the disruption caused by frequent rescheduling. Robust optimization over time (ROOT) provides a useful perspective for BACAP because repeated schedule adjustments may disrupt berth plans, QC assignments, and service continuity~\cite{yazdani2023robust}.

Future research can investigate ROOT-oriented formulations for BACAP by jointly considering long-term robustness, rescheduling cost, and operational stability. This includes integrating rolling-horizon adaptation with data-driven prediction, adjusting robustness levels over time, and designing recourse-aware search operators that preserve robust historical patterns while allowing smooth solution transitions. This direction extends robustness from static ex-ante protection to whole-horizon robustness maintenance.

\subsubsection{Non-stationary Uncertainty}

In practical port operations, uncertainty rarely remains stationary~\cite{wang2024robust}. Vessel arrival patterns, weather conditions, operational efficiency, and tidal constraints may evolve over time, making it difficult to characterize uncertainty using fixed probability distributions or static uncertainty sets. Once the operational environment changes, the robustness of a schedule may deteriorate because historical data no longer represent future uncertainty dynamics~\cite{wang2024proactive}.

Future research can develop robust optimization frameworks that explicitly handle non-stationary uncertainty in dynamically evolving port systems. One direction is adaptive uncertainty set learning, where uncertainty representations are updated using streaming operational data rather than fixed intervals or budgets~\cite{tang2023distributionally}. By integrating change-point detection and online learning into population-based metaheuristics~\cite{hartland2007change}, uncertainty sets associated with vessel delays and handling times can adapt to changing operational conditions.

Another direction is time-adaptive robustness. Existing robust BACAP approaches often apply a uniform robustness level over the planning horizon, although different operational stages may require different levels of protection. Future metaheuristic frameworks can use dynamically adjustable robustness criteria to balance conservativeness and flexibility according to the current operational state and planning horizon.

\subsection{Conclusion}

This paper reviewed robust population-based metaheuristics for BACAP under uncertainty and organized existing studies from a mechanism-oriented perspective. The review summarized uncertainty sources and information representations, solution representation and decoding, robust evaluation and selection, robustness-guided search dynamics, and feasibility preservation and recovery. To support controlled empirical comparison, this paper also presented a benchmark suite with representative uncertainty scenarios and reported illustrative baseline results using GA, ACO, and PSO combined with MMRO, ERO, and TSRO. The baseline results provide initial evidence on how population-based search mechanisms and robustness strategies interact under uncertain BACAP settings. Overall, this review highlights the need for benchmark extension and real-world validation, robustness-aware search operators, time-adaptive robustness, and non-stationary uncertainty handling, which are important for future research on uncertain BACAP and robust port-terminal scheduling.
\bibliographystyle{IEEEtran}
\bibliography{ref}

\end{document}


\title{Supplementary Document for ``Robust Metaheuristics under Uncertainty for Berth Allocation and Quay Crane Assignment: A Review''}
	
	\maketitle

	This supplementary document provides the reproducibility details and additional experimental results that are not included in the main text. It first explains the evaluation metrics used in the benchmark experiments and then reports the additional baseline results for the 50-vessel and 200-vessel settings.
	
	\section{Supplementary Reproducibility Protocol for Evaluation}
	
	This section provides the reproducible evaluation protocol used in the benchmark experiments. The notation follows the main paper. For vessel \(i\in\mathcal{V}\), \(a_i\) denotes the nominal arrival time, \(s_i\) denotes the service start time, \(d_i\) denotes the completion time, and \(l_i\) denotes the vessel length.
	
	\subsection{Reproducible Evaluation Metrics}
	
	The benchmark uses two indicators: average port time and robustness index. The average port time follows common performance criteria in berth allocation and quay-crane assignment studies, where vessel waiting time, service duration, and delay-related measures are widely used to assess operational efficiency~\cite{bierwirth2010survey,bierwirth2015follow,imai2005berth}. The robustness index is designed as a slack-based survival measure, motivated by robust berth scheduling studies in which buffer time, slack, and temporal separation are used to absorb vessel-arrival and handling-time uncertainty and reduce delay propagation~\cite{iris2019recoverable}. Berth-space overlap and QC-sharing conflicts are evaluated according to the standard space--time conflict logic of continuous berth allocation and quay-crane assignment models~\cite{malekahmadi2020integrated,himmiche2023robustness}.
	
	\subsubsection{Average Port Time}
	
	For vessel \(i\), the nominal service duration is \(d_i-s_i\). Let \(q_i^{\mathrm{plan}}\) denote the planned number of QCs assigned to vessel \(i\), and let \(q_i^{\mathrm{real}}\) denote the number of available QCs among those assigned to vessel \(i\) under the realized QC availability state. The adjusted service duration is computed as
	\begin{equation}
		D_i =
		\begin{cases}
			(d_i-s_i)\dfrac{q_i^{\mathrm{plan}}}{q_i^{\mathrm{real}}},
			& q_i^{\mathrm{real}}>0,\\[0.8em]
			\kappa(d_i-s_i),
			& q_i^{\mathrm{real}}=0,
		\end{cases}
	\end{equation}
	where \(\kappa>1\) is a penalty coefficient for complete QC unavailability. This adjustment reflects the dependence of handling duration on QC availability and service capacity, which is consistent with integrated berth allocation and quay-crane scheduling models where reduced crane availability prolongs vessel service time~\cite{imai2008simultaneous,meisel2009heuristics,li2015solving}.
	
	Let \(W_i^{\mathrm{tide}}\) denote the tidal waiting time of vessel \(i\). If \(s_i\) lies within an admissible tidal window, then \(W_i^{\mathrm{tide}}=0\). Otherwise, \(W_i^{\mathrm{tide}}\) is computed as the waiting time from \(s_i\) to the beginning of the next admissible tidal window. The tidal waiting definition follows berth scheduling models with tidal-window constraints, where service feasibility depends on navigational or tidal accessibility~\cite{zhen2017daily}.
	
	The arrival-related service mismatch is defined as
	\begin{equation}
		W_i^{\mathrm{arr}}=|a_i-s_i|.
		\label{eq:supp_arrival_deviation}
	\end{equation}
	This term measures the service-start deviation from the nominal arrival time and is used as an arrival-related waiting or mismatch component in the benchmark evaluation.
	
	The evaluated port time of vessel \(i\) is then
	\begin{equation}
		T_i^{\mathrm{port}}=D_i+W_i^{\mathrm{tide}}+W_i^{\mathrm{arr}}.
	\end{equation}
	The average port time is computed as
	\begin{equation}
		\bar{T}_{\mathrm{port}}=\frac{1}{|\mathcal{V}|}\sum_{i\in\mathcal{V}}T_i^{\mathrm{port}}.
	\end{equation}
	A smaller \(\bar{T}_{\mathrm{port}}\) indicates better operational efficiency.
	
	\subsubsection{Robustness Index}
	
	Robustness is evaluated through a slack-based survival measure. For vessel \(i\), let \(D_i\) denote the QC-availability-adjusted service duration. The adjusted completion time under realized QC availability is
	\begin{equation}
		\tilde d_i=s_i+D_i.
		\label{eq:supp_adjusted_completion}
	\end{equation}
	
	The survival time of vessel \(i\) is defined as
	\begin{equation}
		\sigma_i
		=
		R_i^{\mathrm{tide}}
		+
		R_i^{\mathrm{berth}}
		+
		R_i^{\mathrm{QC}},
		\label{eq:supp_survival_time}
	\end{equation}
	where \(R_i^{\mathrm{tide}}\) denotes the tidal buffer, \(R_i^{\mathrm{berth}}\) denotes the berth-related conflict margin, and \(R_i^{\mathrm{QC}}\) denotes the QC-related conflict margin.
	
	The tidal buffer is computed as
	\begin{equation}
		R_i^{\mathrm{tide}}=\max\{\overline{t}_g-\tilde d_i,0\},
	\end{equation}
	if \(s_i\) lies within tidal window \([\underline{t}_g,\overline{t}_g]\). If \(s_i\) does not lie within any admissible tidal window, then \(R_i^{\mathrm{tide}}=0\).
	
	The berth-conflict term considers vessels whose berth intervals overlap with vessel \(i\). Let the berth interval of vessel \(i\) be \([x_i,x_i+l_i]\). For any vessel \(j\neq i\), a berth-space overlap exists if
	\begin{equation}
		x_i<x_j+l_j
		\quad\text{and}\quad
		x_i+l_i>x_j.
	\end{equation}
	Let \(\mathcal{B}_i\) denote the set of vessels that overlap with vessel \(i\) in berth space and start service before the adjusted completion time of vessel \(i\):
	\begin{equation}
		\mathcal{B}_i
		=
		\left\{
		j\in\mathcal{V}\setminus\{i\}:
		x_i<x_j+l_j,\;
		x_i+l_i>x_j,\;
		s_i\le s_j<\tilde d_i
		\right\}.
	\end{equation}
	For each vessel \(j\in\mathcal{B}_i\), the berth-conflict gap is
	\begin{equation}
		G_{ij}^{\mathrm{berth}}=s_j - \tilde d_i.
	\end{equation}
	The berth-related conflict margin is
	\begin{equation}
		R_i^{\mathrm{berth}}
		=
		\begin{cases}
			\min\limits_{j\in\mathcal{B}_i}G_{ij}^{\mathrm{berth}},
			& \mathcal{B}_i\neq\emptyset,\\[0.8em]
			0,
			& \mathcal{B}_i=\emptyset.
		\end{cases}
	\end{equation}
	
	The QC-conflict term considers vessels sharing at least one assigned QC with vessel \(i\). Let \(\mathcal{Q}_i\) denote the set of QCs assigned to vessel \(i\). The set of vessels sharing at least one QC with vessel \(i\) is
	\begin{equation}
		\mathcal{C}_i=
		\left\{
		j\in\mathcal{V}\setminus\{i\}:
		\mathcal{Q}_i\cap\mathcal{Q}_j\neq\emptyset,\;
		s_i\le s_j<\tilde d_i
		\right\}.
	\end{equation}
	For each vessel \(j\in\mathcal{C}_i\), the QC-conflict gap is
	\begin{equation}
		G_{ij}^{\mathrm{QC}}=s_j - \tilde d_i.
	\end{equation}
	The QC-related conflict margin is
	\begin{equation}
		R_i^{\mathrm{QC}}
		=
		\begin{cases}
			\min\limits_{j\in\mathcal{C}_i}G_{ij}^{\mathrm{QC}},
			& \mathcal{C}_i\neq\emptyset,\\[0.8em]
			0,
			& \mathcal{C}_i=\emptyset.
		\end{cases}
	\end{equation}
	Negative values of \(R_i^{\mathrm{berth}}\) or \(R_i^{\mathrm{QC}}\) indicate detected temporal conflicts or insufficient separation. A value of zero indicates that no corresponding conflict is detected under the defined conflict set.
	
	The survival threshold is defined as the empirical median of the nominal service durations:
	\begin{equation}
		\theta=P_{50}\left(\{d_i-s_i:i\in\mathcal{V}\}\right).
	\end{equation}
	The set of robustness-effective vessels is
	\begin{equation}
		S=\{i\in\mathcal{V}:\sigma_i>\theta\}.
	\end{equation}
	The robustness index is
	\begin{equation}
		\bar{\sigma}
		=
		\begin{cases}
			\displaystyle \frac{1}{|S|}\sum\limits_{i\in S}\sigma_i,
			& |S|>0,\\[0.8em]
			0,
			& |S|=0.
		\end{cases}
		\label{eq:supp_ri}
	\end{equation}
	A larger \(\bar{\sigma}\) indicates stronger slack-based robustness.
	
	\section{Supplementary Experiments}
	\begin{table*}[htbp]\centering
		\scriptsize
		\caption{Additional baseline results (mean $\pm$ std) of average port time and robustness index for nine robust population-based metaheuristics under the 50-vessel setting.}
		\label{tab:50_A}
		\resizebox{\textwidth}{!}{%
			\begin{tabular}{lllccccccccc}
				\toprule
				Cases & Metric & Value& GA-MMRO & GA-ERO & GA-TSRO & ACO-MMRO & ACO-ERO & ACO-TSRO & PSO-MMRO & PSO-ERO & PSO-TSRO \\
				\midrule
				
				C1 & $\bar{T}_{\mathrm{port}}$ &Mean &1.79E+01 & 1.78E+01 & 1.79E+01 & 1.72E+01 & 1.53E+01 & 1.50E+01 & 1.74E+01 & \textbf{1.39E+01} & 1.42E+01 \\
				&              & Std&7.30E-01 & 9.00E-01 & 7.40E-01 & 8.50E-01 & 6.50E-01 & 6.60E-01 & 8.70E-01 & \textbf{8.80E-01} & 7.80E-01 \\
				& $\bar{\sigma}$   & Mean &6.78E+00 & 6.92E+00 & 7.04E+00 & 7.14E+00 & 7.00E+00 & 7.13E+00 & 7.03E+00 & \textbf{7.30E+00} & 7.25E+00 \\
				&              & Std&4.30E-01 & 5.20E-01 & 5.30E-01 & 4.70E-01 & 5.60E-01 & 5.10E-01 & 4.90E-01 & \textbf{5.70E-01} & 5.30E-01 \\
				C2 & $\bar{T}_{\mathrm{port}}$ & Mean &1.80E+01 & 1.80E+01 & 1.82E+01 & 1.76E+01 & 1.51E+01 & 1.50E+01 & 1.76E+01 & 1.42E+01 & \textbf{1.42E+01} \\
				&              & Std&8.30E-01 & 7.00E-01 & 8.40E-01 & 7.00E-01 & 5.40E-01 & 7.60E-01 & 9.10E-01 & 1.05E+00 & \textbf{8.50E-01} \\
				& $\bar{\sigma}$ & Mean &6.86E+00 & 6.63E+00 & 6.91E+00 & 7.03E+00 & \textbf{7.30E+00} & 7.25E+00 & 7.08E+00 & 7.12E+00 & 7.09E+00 \\
				&              & Std&3.70E-01 & 3.60E-01 & 4.90E-01 & 3.40E-01 & \textbf{4.40E-01} & 5.10E-01 & 4.80E-01 & 4.20E-01 & 5.60E-01 \\
				C3 & $\bar{T}_{\mathrm{port}}$ & Mean &2.36E+01 & 2.35E+01 & 2.32E+01 & 2.33E+01 & 2.17E+01 & 2.18E+01 & 2.34E+01 & 2.11E+01 & \textbf{2.11E+01} \\
				&              & Std&9.10E-01 & 7.90E-01 & 5.90E-01 & 8.80E-01 & 7.40E-01 & 6.80E-01 & 8.10E-01 & 1.02E+00 & \textbf{7.40E-01} \\
				& $\bar{\sigma}$ & Mean &6.93E+00 & 6.85E+00 & 6.90E+00 & 7.01E+00 & 6.94E+00 & 6.88E+00 & \textbf{7.03E+00} & 6.40E+00 & 6.44E+00 \\
				&              & Std&5.50E-01 & 4.20E-01 & 4.80E-01 & 5.60E-01 & 4.90E-01 & 5.90E-01 & \textbf{4.60E-01} & 6.10E-01 & 4.40E-01 \\
				C4 & $\bar{T}_{\mathrm{port}}$ & Mean &2.11E+01 & 2.11E+01 & 2.10E+01 & 2.18E+01 & \textbf{2.02E+01} & 2.07E+01 & 2.15E+01 & 2.09E+01 & 2.08E+01 \\
				&              & Std&5.60E-01 & 5.40E-01 & 6.20E-01 & 6.20E-01 & \textbf{6.70E-01} & 8.50E-01 & 7.70E-01 & 8.10E-01 & 7.00E-01 \\
				& $\bar{\sigma}$ &  Mean &6.93E+00 & 7.00E+00 & 6.80E+00 & 6.98E+00 & 7.01E+00 & 6.95E+00 & \textbf{7.03E+00} & 6.88E+00 & 6.97E+00 \\
				&              & Std&6.00E-01 & 4.70E-01 & 4.20E-01 & 5.20E-01 & 5.90E-01 & 5.30E-01 & \textbf{5.20E-01} & 4.60E-01 & 4.50E-01 \\
				\midrule
				
				C5 & $\bar{T}_{\mathrm{port}}$ & Mean &1.98E+01 & 1.97E+01 & 1.98E+01 & 1.90E+01 & 1.69E+01 & 1.66E+01 & 1.92E+01 & \textbf{1.50E+01} & 1.53E+01 \\
				&              & Std&8.00E-01 & 9.60E-01 & 7.70E-01 & 8.70E-01 & 7.00E-01 & 6.60E-01 & 8.80E-01 & \textbf{9.60E-01} & 8.10E-01 \\
				& $\bar{\sigma}$ & Mean &7.00E+00 & 7.13E+00 & 7.19E+00 & 7.36E+00 & 7.12E+00 & 7.24E+00 & 7.13E+00 & \textbf{7.46E+00} & 7.31E+00 \\
				&              & Std&4.40E-01 & 4.50E-01 & 4.80E-01 & 5.10E-01 & 5.70E-01 & 5.60E-01 & 4.60E-01 & \textbf{4.40E-01} & 4.50E-01 \\
				C6 &$\bar{T}_{\mathrm{port}}$ & Mean &1.99E+01 & 1.98E+01 & 2.00E+01 & 1.93E+01 & 1.67E+01 & 1.66E+01 & 1.94E+01 & \textbf{1.53E+01} & 1.54E+01 \\
				&              & Std&8.50E-01 & 7.30E-01 & 8.70E-01 & 7.20E-01 & 5.60E-01 & 8.10E-01 & 9.60E-01 & \textbf{1.10E+00} & 8.50E-01 \\
				& $\bar{\sigma}$ & Mean &6.98E+00 & 6.88E+00 & 7.14E+00 & 7.17E+00 & 7.36E+00 & \textbf{7.51E+00} & 7.26E+00 & 7.43E+00 & 7.22E+00 \\
				&              & Std&4.50E-01 & 3.40E-01 & 5.80E-01 & 4.40E-01 & 4.10E-01 & \textbf{5.10E-01} & 5.20E-01 & 5.10E-01 & 4.80E-01 \\
				C7 &$\bar{T}_{\mathrm{port}}$ & Mean &2.51E+01 & 2.51E+01 & 2.48E+01 & 2.48E+01 & 2.30E+01 & 2.32E+01 & 2.49E+01 & 2.24E+01 & \textbf{2.24E+01} \\
				&              & Std&9.40E-01 & 8.50E-01 & 5.40E-01 & 8.70E-01 & 7.50E-01 & 7.10E-01 & 8.40E-01 & 1.01E+00 & \textbf{7.70E-01} \\
				&$\bar{\sigma}$& Mean &7.05E+00 & 7.12E+00 & 7.10E+00 & 7.13E+00 & 7.07E+00 & 6.98E+00 & \textbf{7.18E+00} & 6.54E+00 & 6.52E+00 \\
				&              & Std&6.10E-01 & 4.60E-01 & 5.40E-01 & 6.30E-01 & 4.90E-01 & 4.80E-01 & \textbf{5.20E-01} & 5.40E-01 & 4.70E-01 \\
				C8 & $\bar{T}_{\mathrm{port}}$ & Mean &2.15E+01 & 2.15E+01 & 2.13E+01 & 2.20E+01 & \textbf{2.01E+01} & 2.05E+01 & 2.17E+01 & 2.05E+01 & 2.04E+01 \\
				&              & Std&6.00E-01 & 6.50E-01 & 6.80E-01 & 6.30E-01 & \textbf{6.00E-01} & 8.70E-01 & 7.00E-01 & 7.50E-01 & 7.00E-01 \\
				& $\bar{\sigma}$ & Mean &7.12E+00 & 7.17E+00 & 6.98E+00 & 7.08E+00 & 7.10E+00 & 7.13E+00 & \textbf{7.21E+00} & 7.09E+00 & 7.09E+00 \\
				&              & Std&6.60E-01 & 4.80E-01 & 4.00E-01 & 5.00E-01 & 5.20E-01 & 5.30E-01 & \textbf{5.10E-01} & 5.10E-01 & 3.80E-01 \\
				\midrule
				
				C9 & $\bar{T}_{\mathrm{port}}$ & Mean &2.15E+01 & 2.13E+01 & 2.15E+01 & 2.07E+01 & 1.88E+01 & 1.85E+01 & 2.11E+01 & \textbf{1.68E+01} & 1.71E+01 \\
				&              & Std&8.90E-01 & 9.60E-01 & 8.20E-01 & 9.00E-01 & 8.20E-01 & 6.90E-01 & 9.30E-01 & \textbf{1.04E+00} & 8.60E-01 \\
				& $\bar{\sigma}$ & Mean &6.62E+00 & 6.75E+00 & 6.86E+00 & 6.94E+00 & 6.72E+00 & 6.85E+00 & 6.89E+00 & 6.97E+00 & \textbf{7.03E+00} \\
				&              & Std&4.20E-01 & 5.40E-01 & 5.70E-01 & 5.50E-01 & 5.40E-01 & 5.30E-01 & 5.00E-01 & 6.50E-01 & \textbf{5.10E-01} \\
				C10 & $\bar{T}_{\mathrm{port}}$ & Mean &2.16E+01 & 2.18E+01 & 2.18E+01 & 2.11E+01 & 1.85E+01 & 1.84E+01 & 2.13E+01 & 1.71E+01 & \textbf{1.71E+01} \\
				&              & Std&1.00E+00 & 8.10E-01 & 9.80E-01 & 8.20E-01 & 7.00E-01 & 9.20E-01 & 1.04E+00 & 1.13E+00 & \textbf{9.70E-01} \\
				& $\bar{\sigma}$ & Mean &6.87E+00 & 6.87E+00 & 6.87E+00 & 6.73E+00 & \textbf{7.16E+00} & 7.12E+00 & 6.86E+00 & 7.05E+00 & 6.90E+00 \\
				&              & Std&5.20E-01 & 4.60E-01 & 5.70E-01 & 4.10E-01 & \textbf{5.30E-01} & 4.90E-01 & 6.50E-01 & 6.30E-01 & 3.90E-01 \\
				C11 & $\bar{T}_{\mathrm{port}}$ & Mean &2.69E+01 & 2.68E+01 & 2.65E+01 & 2.66E+01 & 2.47E+01 & 2.50E+01 & 2.66E+01 & 2.42E+01 & \textbf{2.41E+01} \\
				&              & Std&1.00E+00 & 9.50E-01 & 5.90E-01 & 8.80E-01 & 7.10E-01 & 7.30E-01 & 9.30E-01 & 1.03E+00 & \textbf{9.00E-01} \\
				& $\bar{\sigma}$ & Mean &7.02E+00 & 7.13E+00 & \textbf{7.21E+00} & 6.97E+00 & 6.86E+00 & 6.83E+00 & 7.03E+00 & 6.77E+00 & 6.83E+00 \\
				&              & Std&5.90E-01 & 5.20E-01 & \textbf{5.80E-01} & 4.50E-01 & 5.90E-01 & 6.90E-01 & 5.70E-01 & 6.20E-01 & 7.00E-01 \\
				C12 & $\bar{T}_{\mathrm{port}}$ & Mean &2.32E+01 & 2.31E+01 & 2.30E+01 & 2.38E+01 & \textbf{2.18E+01} & 2.23E+01 & 2.35E+01 & 2.23E+01 & 2.22E+01 \\
				&              & Std&6.50E-01 & 7.60E-01 & 8.00E-01 & 6.80E-01 & \textbf{6.80E-01} & 9.30E-01 & 7.70E-01 & 8.00E-01 & 7.30E-01 \\
				& $\bar{\sigma}$ & Mean &6.99E+00 & 6.90E+00 & 6.82E+00 & 6.89E+00 & 6.98E+00 & 7.01E+00 & 7.04E+00 & \textbf{7.06E+00} & 7.02E+00 \\
				&              & Std&6.60E-01 & 5.30E-01 & 4.80E-01 & 5.30E-01 & 4.90E-01 & 5.50E-01 & 5.50E-01 & \textbf{5.60E-01} & 5.30E-01 \\
				\midrule
				
				C13 & $\bar{T}_{\mathrm{port}}$ & Mean &3.01E+01 & 3.00E+01 & 3.03E+01 & 2.87E+01 & 2.80E+01 & 2.70E+01 & 2.91E+01 & \textbf{2.53E+01} & 2.57E+01 \\
				&              & Std&1.10E+00 & 1.00E+00 & 9.90E-01 & 9.50E-01 & 9.80E-01 & 8.70E-01 & 9.80E-01 & \textbf{1.15E+00} & 9.40E-01 \\
				& $\bar{\sigma}$ & Mean &7.42E+00 & 7.32E+00 & 7.35E+00 & 7.46E+00 & 7.21E+00 & 7.34E+00 & 7.22E+00 & \textbf{7.51E+00} & 7.40E+00 \\
				&              & Std&4.90E-01 & 5.00E-01 & 4.80E-01 & 5.50E-01 & 5.10E-01 & 5.30E-01 & 5.40E-01 & \textbf{5.80E-01} & 4.80E-01 \\
				C14 & $\bar{T}_{\mathrm{port}}$ & Mean &2.98E+01 & 2.98E+01 & 3.00E+01 & 2.84E+01 & 2.72E+01 & 2.66E+01 & 2.91E+01 & \textbf{2.57E+01} & 2.58E+01 \\
				&              & Std&1.04E+00 & 9.10E-01 & 1.02E+00 & 8.90E-01 & 8.60E-01 & 9.40E-01 & 1.07E+00 & \textbf{1.17E+00} & 9.50E-01 \\
				& $\bar{\sigma}$ & Mean &7.44E+00 & 7.27E+00 & 7.34E+00 & 7.41E+00 & 7.33E+00 & 7.36E+00 & 7.24E+00 & \textbf{7.52E+00} & 7.45E+00 \\
				&              & Std&4.50E-01 & 4.80E-01 & 5.10E-01 & 5.20E-01 & 5.30E-01 & 5.30E-01 & 5.20E-01 & \textbf{5.40E-01} & 4.80E-01 \\
				C15 & $\bar{T}_{\mathrm{port}}$ & Mean &3.68E+01 & 3.65E+01 & 3.62E+01 & 3.56E+01 & 3.37E+01 & 3.33E+01 & 3.57E+01 & \textbf{3.24E+01} & 3.26E+01 \\
				&              & Std&1.11E+00 & 9.60E-01 & 6.40E-01 & 1.00E+00 & 8.70E-01 & 9.00E-01 & 1.00E+00 & \textbf{1.05E+00} & 8.90E-01 \\
				& $\bar{\sigma}$ & Mean &7.17E+00 & 7.21E+00 & \textbf{7.30E+00} & 7.21E+00 & 7.15E+00 & 7.16E+00 & 7.27E+00 & 6.91E+00 & 6.95E+00 \\
				&              & Std&5.30E-01 & 5.10E-01 & \textbf{5.70E-01} & 6.10E-01 & 5.30E-01 & 6.20E-01 & 6.10E-01 & 5.90E-01 & 6.20E-01 \\
				C16 & $\bar{T}_{\mathrm{port}}$ & Mean &3.13E+01 & 3.11E+01 & 3.09E+01 & 3.18E+01 & 3.02E+01 & 3.06E+01 & 3.14E+01 & \textbf{2.98E+01} & 3.03E+01 \\
				&              & Std&7.90E-01 & 9.50E-01 & 1.00E+00 & 7.60E-01 & 6.90E-01 & 9.40E-01 & 8.30E-01 & \textbf{9.50E-01} & 7.90E-01 \\
				& $\bar{\sigma}$ & Mean &6.94E+00 & 7.03E+00 & 6.91E+00 & 6.91E+00 & 7.05E+00 & 6.97E+00 & 7.08E+00 & \textbf{7.12E+00} & 6.98E+00 \\
				&              & Std&5.00E-01 & 5.50E-01 & 4.50E-01 & 5.20E-01 & 4.40E-01 & 5.10E-01 & 5.40E-01 & \textbf{5.90E-01} & 5.00E-01 \\
				\bottomrule
			\end{tabular}
		}
	\end{table*}
	\begin{table*}[htbp]
		\centering
		\scriptsize
		\caption{Additional baseline results (mean $\pm$ std) of average port time and robustness index for nine robust population-based metaheuristics under the 200-vessel setting.}
		\label{tab:200_A}
		\resizebox{\textwidth}{!}{
			\begin{tabular}{lllccccccccc}
				\toprule
				Cases & Metric & Value & GA-MMRO & GA-ERO & GA-TSRO & ACO-MMRO & ACO-ERO & ACO-TSRO & PSO-MMRO & PSO-ERO & PSO-TSRO \\
				\midrule
				
				C1 &$\bar{T}_{\mathrm{port}}$& Mean& 5.75E+01 & 5.77E+01 & 5.74E+01 & 5.68E+01 & 5.36E+01 & 5.37E+01 & 5.69E+01 & \textbf{5.27E+01} & 5.27E+01 \\
				&              & Std&6.50E-01 & 1.13E+00 & 1.07E+00 & 8.60E-01 & 9.20E-01 & 9.00E-01 & 1.05E+00 & \textbf{1.04E+00} & 1.16E+00 \\
				& $\bar{\sigma}$ & Mean&6.96E+00 & 6.99E+00 & \textbf{7.11E+00} & 7.00E+00 & 7.08E+00 & 7.02E+00 & 7.01E+00 & 7.07E+00 & 7.09E+00 \\
				&              & Std&2.20E-01 & 2.40E-01 & \textbf{1.90E-01} & 2.90E-01 & 2.30E-01 & 2.30E-01 & 2.40E-01 & 2.00E-01 & 2.90E-01 \\
				C2 & $\bar{T}_{\mathrm{port}}$ & Mean&5.75E+01 & 5.70E+01 & 5.70E+01 & 5.69E+01 & 5.34E+01 & 5.33E+01 & 5.69E+01 & \textbf{5.22E+01} & 5.23E+01 \\
				&              & Std&1.06E+00 & 1.25E+00 & 9.50E-01 & 8.80E-01 & 8.60E-01 & 9.00E-01 & 1.03E+00 & \textbf{1.07E+00} & 1.33E+00 \\
				& $\bar{\sigma}$ & Mean&6.97E+00 & 7.01E+00 & 6.96E+00 & 7.06E+00 & 7.02E+00 & 6.99E+00 & 7.07E+00 & \textbf{7.11E+00} & 7.05E+00 \\
				&              & Std&2.50E-01 & 2.40E-01 & 2.30E-01 & 2.10E-01 & 2.50E-01 & 2.30E-01 & 1.90E-01 & \textbf{2.90E-01} & 2.00E-01 \\
				C3 & $\bar{T}_{\mathrm{port}}$ & Mean&5.86E+01 & 5.88E+01 & 5.89E+01 & 5.82E+01 & 5.65E+01 & 5.58E+01 & 5.86E+01 & 5.49E+01 & \textbf{5.47E+01} \\
				&              & Std&9.50E-01 & 1.08E+00 & 1.01E+00 & 1.07E+00 & 9.30E-01 & 9.90E-01 & 1.16E+00 & 8.60E-01 & \textbf{1.10E+00} \\
				& $\bar{\sigma}$ & Mean&6.99E+00 & 6.94E+00 & 6.95E+00 & 7.01E+00 & \textbf{7.07E+00} & 6.85E+00 & 6.97E+00 & 6.99E+00 & 6.93E+00 \\
				&              & Std&3.00E-01 & 2.40E-01 & 2.10E-01 & 2.50E-01 & \textbf{2.80E-01} & 1.90E-01 & 2.50E-01 & 2.80E-01 & 2.70E-01 \\
				C4 & $\bar{T}_{\mathrm{port}}$ & Mean&5.45E+01 & 5.47E+01 & 5.47E+01 & 5.50E+01 & 5.28E+01 & 5.26E+01 & 5.46E+01 & 5.22E+01 & \textbf{5.21E+01} \\
				&              & Std&1.05E+00 & 7.10E-01 & 9.50E-01 & 9.00E-01 & 9.00E-01 & 8.80E-01 & 1.09E+00 & 9.70E-01 & \textbf{9.00E-01} \\
				& $\bar{\sigma}$ & Mean&7.03E+00 & 7.00E+00 & 7.08E+00 & 6.98E+00 & 7.02E+00 & 7.05E+00 & \textbf{7.09E+00} & 6.95E+00 & 7.06E+00 \\
				&              & Std&2.70E-01 & 2.50E-01 & 2.80E-01 & 2.50E-01 & 3.00E-01 & 2.40E-01 & \textbf{2.40E-01} & 3.10E-01 & 2.90E-01 \\
				\midrule
				
				C5 & $\bar{T}_{\mathrm{port}}$ & Mean&5.91E+01 & 5.93E+01 & 5.90E+01 & 5.84E+01 & 5.51E+01 & 5.52E+01 & 5.84E+01 & \textbf{5.41E+01} & 5.41E+01 \\
				&              & Std&6.50E-01 & 1.14E+00 & 1.07E+00 & 8.60E-01 & 9.30E-01 & 9.10E-01 & 1.06E+00 & \textbf{1.05E+00} & 1.17E+00 \\
				& $\bar{\sigma}$ & Mean&7.23E+00 & 7.22E+00 & 7.36E+00 & 7.32E+00 & \textbf{7.37E+00} & 7.30E+00 & 7.26E+00 & 7.33E+00 & 7.35E+00 \\
				&              & Std&2.40E-01 & 2.20E-01 & 2.10E-01 & 3.00E-01 & \textbf{2.50E-01} & 2.20E-01 & 2.90E-01 & 2.40E-01 & 2.90E-01 \\
				C6 & $\bar{T}_{\mathrm{port}}$ & Mean&5.91E+01 & 5.86E+01 & 5.85E+01 & 5.84E+01 & 5.49E+01 & 5.48E+01 & 5.84E+01 & \textbf{5.36E+01} & 5.37E+01 \\
				&              & Std&1.06E+00 & 1.24E+00 & 9.60E-01 & 8.80E-01 & 8.60E-01 & 9.10E-01 & 1.03E+00 & \textbf{1.09E+00} & 1.35E+00 \\
				& $\bar{\sigma}$ & Mean&7.20E+00 & 7.32E+00 & 7.25E+00 & 7.28E+00 & 7.29E+00 & 7.27E+00 & \textbf{7.37E+00} & 7.34E+00 & 7.28E+00 \\
				&              & Std&3.10E-01 & 2.40E-01 & 2.00E-01 & 2.50E-01 & 2.60E-01 & 2.20E-01 & \textbf{1.90E-01} & 3.00E-01 & 2.50E-01 \\
				C7 & $\bar{T}_{\mathrm{port}}$ & Mean&6.00E+01 & 6.02E+01 & 6.03E+01 & 5.96E+01 & 5.79E+01 & 5.72E+01 & 5.99E+01 & 5.63E+01 & \textbf{5.60E+01} \\
				&              & Std&9.60E-01 & 1.08E+00 & 1.01E+00 & 1.07E+00 & 9.50E-01 & 1.00E+00 & 1.18E+00 & 8.60E-01 & \textbf{1.11E+00} \\
				& $\bar{\sigma}$ & Mean&7.26E+00 & 7.25E+00 & 7.29E+00 & 7.33E+00 & \textbf{7.34E+00} & 7.12E+00 & 7.28E+00 & 7.27E+00 & 7.24E+00 \\
				&              & Std&2.30E-01 & 2.60E-01 & 2.40E-01 & 3.00E-01 & \textbf{2.00E-01} & 2.20E-01 & 2.30E-01 & 2.90E-01 & 2.80E-01 \\
				C8 & $\bar{T}_{\mathrm{port}}$ & Mean&5.58E+01 & 5.60E+01 & 5.60E+01 & 5.63E+01 & 5.41E+01 & 5.39E+01 & 5.59E+01 & 5.33E+01 & \textbf{5.32E+01} \\
				&              & Std&1.04E+00 & 7.20E-01 & 9.70E-01 & 9.20E-01 & 9.00E-01 & 8.90E-01 & 1.12E+00 & 1.00E+00 & \textbf{9.10E-01} \\
				& $\bar{\sigma}$ & Mean&7.31E+00 & 7.27E+00 & \textbf{7.33E+00} & 7.29E+00 & 7.27E+00 & 7.26E+00 & 7.28E+00 & 7.26E+00 & 7.30E+00 \\
				&              & Std&2.10E-01 & 2.80E-01 & 2.40E-01 & 2.00E-01 & 2.80E-01 & 2.50E-01 & 2.50E-01 & 3.30E-01 & 2.80E-01 \\
				\midrule
				
				C9 & $\bar{T}_{\mathrm{port}}$ & Mean&6.09E+01 & 6.06E+01 & 6.06E+01 & 6.04E+01 & 5.72E+01 & 5.70E+01 & 6.05E+01 & \textbf{5.58E+01} & 5.60E+01 \\
				&              & Std&1.31E+00 & 1.20E+00 & 1.17E+00 & 7.90E-01 & 1.18E+00 & 9.30E-01 & 1.18E+00 & \textbf{1.34E+00} & 1.08E+00 \\
				& $\bar{\sigma}$ & Mean&6.29E+00 & 6.29E+00 & 6.20E+00 & 6.68E+00 & \textbf{6.76E+00} & 6.65E+00 & 6.67E+00 & 6.67E+00 & 6.67E+00 \\
				&              & Std&3.00E-01 & 3.30E-01 & 3.00E-01 & 2.60E-01 & \textbf{2.50E-01} & 2.40E-01 & 2.90E-01 & 2.40E-01 & 1.70E-01 \\
				C10 & $\bar{T}_{\mathrm{port}}$ & Mean&6.08E+01 & 6.06E+01 & 6.06E+01 & 6.03E+01 & 5.66E+01 & 5.65E+01 & 5.99E+01 & 5.58E+01 & \textbf{5.56E+01} \\
				&              & Std&1.09E+00 & 1.01E+00 & 1.00E+00 & 1.34E+00 & 9.70E-01 & 9.10E-01 & 1.10E+00 & 9.10E-01 & \textbf{1.04E+00} \\
				& $\bar{\sigma}$ & Mean&6.23E+00 & 6.19E+00 & 6.18E+00 & 6.70E+00 & \textbf{6.72E+00} & 6.68E+00 & 6.71E+00 & 6.63E+00 & 6.61E+00 \\
				&              & Std&2.80E-01 & 3.20E-01 & 2.70E-01 & 3.00E-01 & \textbf{2.70E-01} & 2.40E-01 & 2.30E-01 & 2.30E-01 & 2.20E-01 \\
				C11 & $\bar{T}_{\mathrm{port}}$ & Mean&6.65E+01 & 6.60E+01 & 6.57E+01 & 6.52E+01 & 6.13E+01 & 6.18E+01 & 6.47E+01 & \textbf{5.98E+01} & 5.99E+01 \\
				&              & Std&8.50E-01 & 8.60E-01 & 1.10E+00 & 7.70E-01 & 7.70E-01 & 9.10E-01 & 1.11E+00 & \textbf{1.04E+00} & 9.30E-01 \\
				& $\bar{\sigma}$ & Mean&6.13E+00 & 6.13E+00 & 6.22E+00 & 6.71E+00 & 6.69E+00 & 6.66E+00 & \textbf{6.75E+00} & 6.65E+00 & 6.75E+00\\
				&              & Std&3.10E-01 & 2.90E-01 & 2.70E-01 & 2.20E-01 & 3.20E-01 & 2.30E-01 & \textbf{2.60E-01} & 2.40E-01 & 2.80E-01 \\
				C12 & $\bar{T}_{\mathrm{port}}$& Mean&6.51E+01 & 6.42E+01 & 6.41E+01 & 6.42E+01 & 6.06E+01 & 5.99E+01 & 6.41E+01 & 5.96E+01 & \textbf{5.93E+01} \\
				&              & Std&1.11E+00 & 9.40E-01 & 8.60E-01 & 1.03E+00 & 9.10E-01 & 9.00E-01 & 1.14E+00 & 8.50E-01 & \textbf{1.25E+00} \\
				& $\bar{\sigma}$ & Mean&6.34E+00 & 6.36E+00 & 6.31E+00 & 6.76E+00 & 6.72E+00 & \textbf{6.80E+00} & 6.66E+00 & 6.72E+00 & 6.74E+00 \\
				&              & Std&2.60E-01 & 3.00E-01 & 3.50E-01 & 2.20E-01 & 2.40E-01 & \textbf{2.50E-01} & 2.40E-01 & 2.60E-01 & 2.30E-01 \\
				\midrule
				
				C13 & $\bar{T}_{\mathrm{port}}$ & Mean&6.00E+01 & 6.02E+01 & 6.00E+01 & 5.93E+01 & 5.62E+01 & 5.62E+01 & 5.93E+01 & \textbf{5.52E+01} & 5.53E+01 \\
				&              & Std&7.00E-01 & 1.09E+00 & 1.02E+00 & 8.40E-01 & 9.30E-01 & 9.30E-01 & 1.02E+00 & \textbf{1.07E+00} & 1.12E+00 \\
				& $\bar{\sigma}$ & Mean&7.11E+00 & 7.09E+00 & 7.23E+00 & 7.14E+00 & 7.20E+00 & \textbf{7.26E+00} & 7.10E+00 & 7.15E+00 & 7.09E+00 \\
				&              & Std&2.30E-01 & 2.70E-01 & 2.70E-01 & 1.90E-01 & 2.20E-01 & \textbf{3.30E-01} & 2.10E-01 & 3.30E-01 & 1.80E-01 \\
				C14 & $\bar{T}_{\mathrm{port}}$ & Mean&6.00E+01 & 5.95E+01 & 5.95E+01 & 5.93E+01 & 5.60E+01 & 5.59E+01 & 5.93E+01 & \textbf{5.47E+01} & 5.48E+01 \\
				&              & Std&1.10E+00 & 1.24E+00 & 1.04E+00 & 9.20E-01 & 9.10E-01 & 8.90E-01 & 1.10E+00 & \textbf{1.12E+00} & 1.35E+00 \\
				& $\bar{\sigma}$ & Mean&7.18E+00 & 7.16E+00 & 7.11E+00 & 7.18E+00 & 7.18E+00 & \textbf{7.29E+00} & 7.25E+00 & 7.22E+00 & 7.17E+00 \\
				&              & Std&2.50E-01 & 3.00E-01 & 2.50E-01 & 2.30E-01 & 2.70E-01 & \textbf{3.00E-01} & 2.40E-01 & 2.20E-01 & 2.90E-01 \\
				C15 & $\bar{T}_{\mathrm{port}}$ & Mean&6.12E+01 & 6.13E+01 & 6.14E+01 & 6.08E+01 & 5.90E+01 & 5.84E+01 & 6.10E+01 & 5.75E+01 & \textbf{5.73E+01} \\
				&              & Std&9.60E-01 & 1.09E+00 & 1.00E+00 & 1.10E+00 & 9.70E-01 & 9.80E-01 & 1.18E+00 & 8.70E-01 & \textbf{1.13E+00} \\
				& $\bar{\sigma}$ & Mean&7.13E+00 & 7.22E+00 & 7.22E+00 & 7.16E+00 & \textbf{7.25E+00} & 7.12E+00 & 7.09E+00 & 7.22E+00 & 7.17E+00 \\
				&              & Std&2.20E-01 & 2.60E-01 & 2.20E-01 & 2.00E-01 & \textbf{3.10E-01} & 2.70E-01 & 3.20E-01 & 2.70E-01 & 2.70E-01 \\
				C16 & $\bar{T}_{\mathrm{port}}$ & Mean&5.70E+01 & 5.72E+01 & 5.71E+01 & 5.75E+01 & 5.53E+01 & 5.51E+01 & 5.71E+01 & 5.48E+01 & \textbf{5.47E+01} \\
				&              & Std&1.03E+00 & 7.60E-01 & 1.03E+00 & 9.60E-01 & 9.30E-01 & 9.00E-01 & 1.11E+00 & 9.00E-01 & \textbf{9.00E-01} \\
				& $\bar{\sigma}$ & Mean&7.18E+00 & 7.17E+00 & 7.13E+00 & 7.15E+00 & 7.19E+00 & \textbf{7.26E+00} & 7.08E+00 & 7.17E+00 & 7.19E+00 \\
				&              & Std&2.70E-01 & 2.40E-01 & 2.40E-01 & 2.70E-01 & 3.10E-01 & \textbf{2.20E-01} & 1.70E-01 & 2.70E-01 & 2.30E-01 \\
				\bottomrule
			\end{tabular}
		}
	\end{table*}

	Table~\ref{tab:50_A} reports the additional baseline results under the 50-vessel setting. In terms of average port time, PSO-based combinations often obtain lower values under the tested cases, especially when combined with ERO or TSRO. ACO-based combinations also show competitive performance in several cases, while GA-based combinations provide useful reference baselines. In terms of robustness, the relative behavior varies across uncertainty scenarios. Some PSO-based combinations achieve high robustness scores in several cases, whereas ACO-MMRO and ACO-TSRO remain competitive under specific uncertainty settings. These results indicate that the interaction between the search mechanism and the robustness strategy affects the efficiency--robustness trade-off even in small-scale instances.
	
	Table~\ref{tab:200_A} reports the additional baseline results under the 200-vessel setting. PSO-based combinations often maintain competitive average port times as the problem scale increases, suggesting that they provide useful efficiency-oriented baselines under the tested settings. ACO-based combinations remain competitive in terms of robustness in several cases, especially when paired with MMRO or TSRO. These observations are consistent with the main text: different metaheuristic frameworks and robustness strategies may favor different parts of the efficiency--robustness trade-off.
	
	Overall, the supplementary results provide additional reference evidence for the proposed benchmark under small- and large-scale settings. They should be interpreted as illustrative baseline results rather than as a definitive ranking of metaheuristic algorithms.
	
	\bibliographystyle{IEEEtran}
	\bibliography{ref1}